\documentclass[conference]{IEEEtran}
\IEEEoverridecommandlockouts
\usepackage{cite}
\usepackage{amsmath,amssymb,amsfonts}
\usepackage{algorithm}
\usepackage{algorithmic}
\usepackage{wrapfig}
\usepackage{transparent}
\usepackage{multicol}
\usepackage{multirow}
\usepackage{subcaption}
\usepackage{float}
\usepackage{stfloats}
\usepackage{graphicx}
\usepackage{textcomp}
\usepackage{xcolor}
\usepackage{comment}
\usepackage{amsmath,amssymb} 
\usepackage{bbm}
\usepackage{xspace}
\usepackage[section]{placeins}
\def\BibTeX{{\rm B\kern-.05em{\sc i\kern-.025em b}\kern-.08em
    T\kern-.1667em\lower.7ex\hbox{E}\kern-.125emX}}
    
\usepackage{url}
\usepackage[pagebackref,breaklinks,colorlinks]{hyperref}
\hypersetup{
 colorlinks=true,
 linkcolor=red,
 filecolor=magenta,      
 urlcolor=magenta,
 citecolor=green
}

\newcommand*{\eg}{\textit{e.g.}\@\xspace}

\newcommand*{\etal}{\textit{et al.}\@\xspace}
\newcommand*{\wrt}{\textit{w.r.t.}\@\xspace}

\begin{document}
\title{Towards Transferable Unrestricted Adversarial Examples with Minimum Changes
}

\author{
\IEEEauthorblockN{Fangcheng Liu}
\IEEEauthorblockA{
Peking University\\
\textit{equation@stu.pku.edu.cn}}
\and
\IEEEauthorblockN{Chao Zhang}
\IEEEauthorblockA{
Peking University\\
\textit{c.zhang@pku.edu.cn}}
\and
\IEEEauthorblockN{Hongyang Zhang}
\IEEEauthorblockA{
University of Waterloo\\
\textit{hongyang.zhang@uwaterloo.ca}}
}

\maketitle

\begin{abstract}
Transfer-based adversarial example is one of the most important classes of black-box attacks. However, there is a trade-off between transferability and imperceptibility of the adversarial perturbation. Prior work in this direction often requires a fixed but large $\ell_p$-norm perturbation budget to reach a good transfer success rate, leading to perceptible adversarial perturbations. On the other hand, most of the current unrestricted adversarial attacks that aim to generate semantic-preserving perturbations suffer from weaker transferability to the target model. In this work, we propose a \emph{geometry-aware framework} to generate transferable adversarial examples with minimum changes. Analogous to model selection in statistical machine learning, we leverage a validation model to select the best perturbation budget for each image under both the $\ell_{\infty}$-norm and unrestricted threat models. We propose a principled method for the partition of training and validation models by encouraging intra-group diversity while penalizing extra-group similarity. Extensive experiments verify the effectiveness of our framework on {balancing} imperceptibility and transferability of the crafted adversarial examples. The methodology is the foundation of our entry to the \emph{CVPR’21 Security AI Challenger: Unrestricted Adversarial Attacks on ImageNet}, in which we ranked 1st place out of 1,559 teams and surpassed the runner-up submissions by 4.59\% and 23.91\% in terms of final score and average image quality level, respectively. Code is available at \url{https://github.com/Equationliu/GA-Attack}.
\end{abstract}


\vspace{-3pt}
\section{Introduction}
\label{sec:intro}
Though deep neural networks have exhibited impressive performance in various fields~\cite{he2016deep,dosovitskiy2021an}, they are vulnerable to adversarial examples~\cite{szegedy2013intriguing,DBLP:journals/corr/GoodfellowSS14,shao2021adversarial,Bhojanapalli_2021_ICCV,bai2021transformers}, where test inputs that have been modified slightly strategically cause misclassification. Adversarial examples have posed serious threats to various security-critical applications, such as autonomous driving~\cite{bojarski2016end} and face recognition~\cite{BMVC2015_41}. Most positive results on adversarial attacks have focused on white-box settings~\cite{Athalye2018,NEURIPS2020_11f38f8e}. However, the problem becomes more challenging when it comes to the black-box setting, where the attacker has no information about the model architecture, hyper-parameters, and even the outputs of the black-box model. In this setting, adversarial examples are typically generated via \emph{transfer-based} methods~\cite{szegedy2013intriguing,papernot2016transferability,papernot2017practical}, \eg, attacking an ensemble of accessible source models and hoping that the same adversarial examples are able to fool the unknown target/test model~\cite{liu2017delving,tram2018}.

\begin{figure*}[t]
  \begin{center}
  \includegraphics[width=1.0\textwidth]{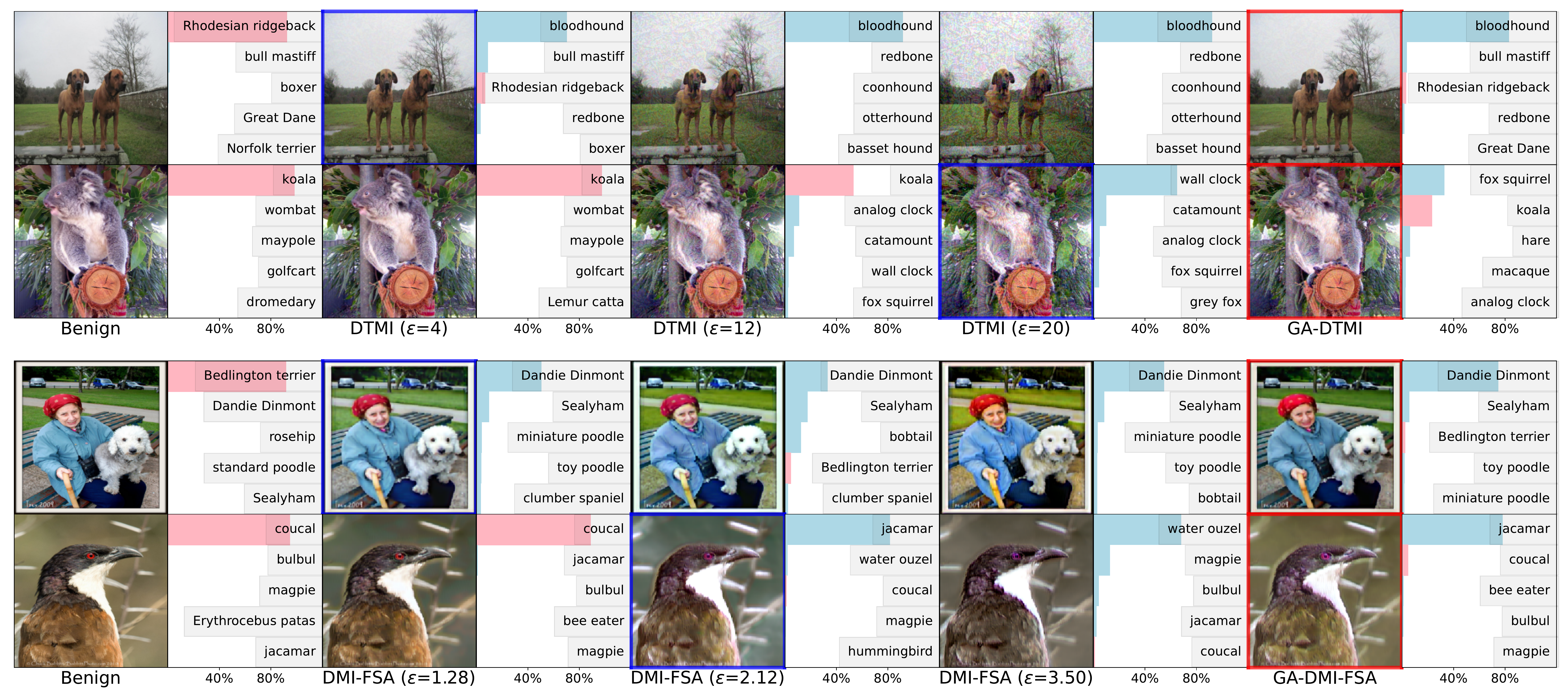}
  \end{center}
  \vspace{-10pt}
  \caption{\textbf{Comparison between our method and two baselines under both the $\ell_{\infty}$-norm (top) and unrestricted (bottom) threat models using various perturbation radii}. In the even columns, we present the top-5 confidence bars of the target model for the images in the left. The ground-truth label is marked by pink and other labels are marked by blue. In each row, the misclassified adversarial example with minimum perturbation radius is highlighted by a blue bounding box, indicating that \emph{the perturbation budgets required for distinct images are different}. Note that the ``human-imperceptible'' constraint is violated when the $\ell_{\infty}$-norm perturbation radius is too large. However, our GA framework generates transferable {unrestricted} adversarial examples (highlighted by red bounding boxes) with lower budgets and smaller changes when compared to the benign images.}
  \label{fig:insight}
  \vspace{-8pt}
\end{figure*}

Despite a large amount of work on transfer-based attacks, many fundamental questions remain unresolved. For example, existing transfer-based attacks~\cite{dong2018boosting,xie2019improving,dong2019evading} that search for adversarial examples in a fixed-radius $\ell_{\infty}$-norm ball often require a high perturbation budget to reach a satisfactory transfer success rate. However, such perturbations might be perceptible to humans (see Figs.~\ref{fig:insight} and \ref{fig:attack_adv}). On the other hand, unrestricted attacks that aim to generate minimum human-imperceptible perturbations~\cite{xiao2018spatially,wong2019wasserstein,laidlaw2021perceptual} suffer from weaker transferability to the target model. This is in part due to the difference between the decision boundaries of the source and target models. {Given the trade-off between transferability and imperceptibility}, one of the long-standing questions is generating transferable adversarial examples by minimum changes of natural examples.

\subsection{Our Methodology and Results}

In this work, we propose a novel geometry-aware framework to generate transferable {unrestricted} adversarial examples with minimum changes. Our intuition is that the smallest perturbation budgets \wrt distinct images should be different (see Fig.~\ref{fig:insight}) and should depend on their geometrical relationship with the decision boundary of the target model (see Fig.~\ref{fig:conceptual}). Unfortunately, finding transferable minimum-budget adversarial perturbations is an intractable optimization problem (see Eq.~\eqref{eq: argmin}) as the target model is unknown. We approximately solve this problem by discretizing the continuous space of perturbation radius into a finite set and choosing the minimum perturbation budget that is able to fool the test model. The main challenge here is to evaluate whether a given perturbation can transfer well to the unknown target model~\cite{NEURIPS2019_32508f53,katzir2021s}.

To overcome this challenge, we split all accessible white-box source models into training and validation sets, where adversarial perturbations are crafted only on the training set. We use the validation set to select the smallest perturbation radius for each input that suffices to fool the validation model with a certain confidence level through an \emph{early-stopping} mechanism. When the training (or validation) set consists of multiple models, we use their average ensemble~\cite{liu2017delving}. Experimentally, our method yields a significant performance boost on the trade-off (leading to higher $S_{total}$ in Table.~\ref{tab:tradeoff}) between transferability and imperceptibility. As shown in Fig.~\ref{fig:score}, the transfer success rate of our method GA-DTMI-FGSM surpasses the baseline DTMI-FGSM (see Eq.~\eqref{DTMI}) by up to 16\% in absolute value under the same average perturbation reward (see Eq.~\eqref{total}). Besides, our method GA-DMI-FSA is able to generate semantic-preserving {yet} transferable unrestricted adversarial examples under the unrestricted threat model (see Eq.~\eqref{unrestricted}, Figs.~\ref{fig:insight}, \ref{fig:attack_adv}, \ref{fig:semantic} and \ref{fig:adv_more}).

\subsection{Summary of Our Contributions}
\label{sec:intro_sum}

\begin{figure*}[h]
  \centering
  \newcommand{\turnheightnew}{0.125\textwidth}
\renewcommand{\arraystretch}{0.0}
\setlength{\tabcolsep}{0pt}
\newcommand{\shiftleft}[2]{\makebox[0pt][r]{\makebox[#1][l]{#2}}}
\newcommand{\imlabel}[2]{\includegraphics[width=\turnheightnew]{#1}%
\raisebox{54pt}{\shiftleft{60pt}{\makebox[-1pt][l]{\small #2}}}}

\centering

\begin{tabular}{cccccccc}
\imlabel{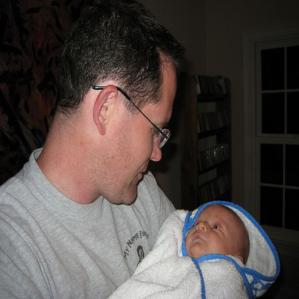}{\textcolor{white}{$\boldsymbol{x}$}} &
\includegraphics[width=\turnheightnew]{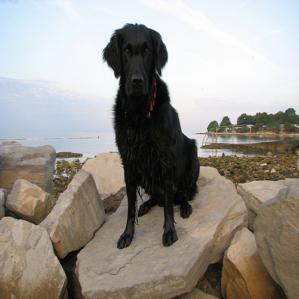} &
\includegraphics[width=\turnheightnew]{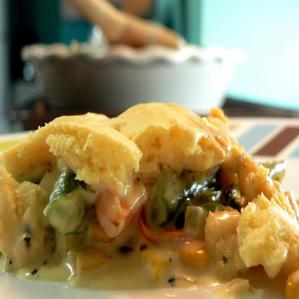} &
\includegraphics[width=\turnheightnew]{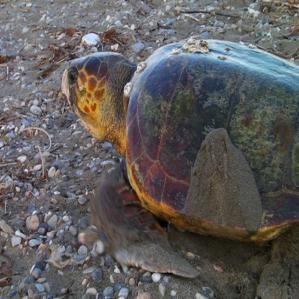} &
\includegraphics[width=\turnheightnew]{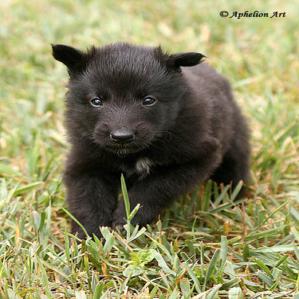} &
\includegraphics[width=\turnheightnew]{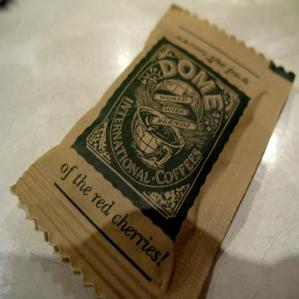} &
\includegraphics[width=\turnheightnew]{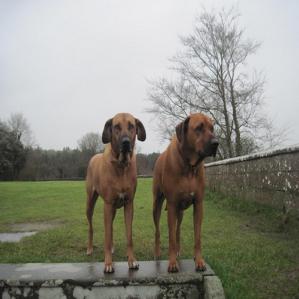} &
\includegraphics[width=\turnheightnew]{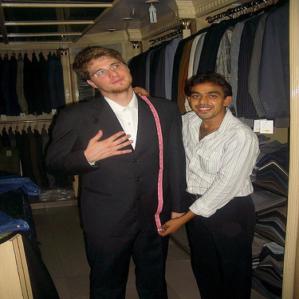} \\ 
\imlabel{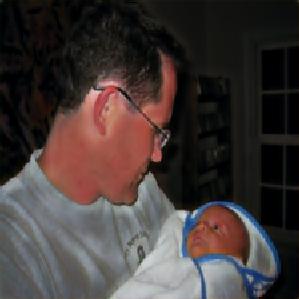}{\textcolor{white}{$\boldsymbol{x}^{\prime}$}} &
\includegraphics[width=\turnheightnew]{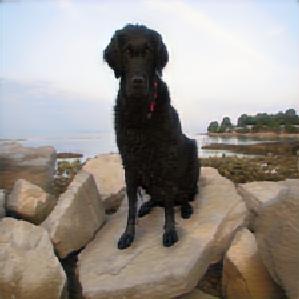} &
\includegraphics[width=\turnheightnew]{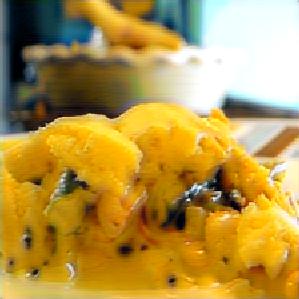} &
\includegraphics[width=\turnheightnew]{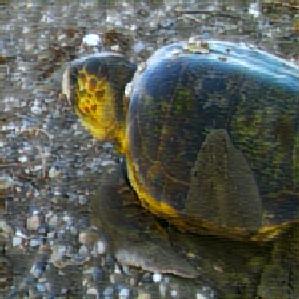} &
\includegraphics[width=\turnheightnew]{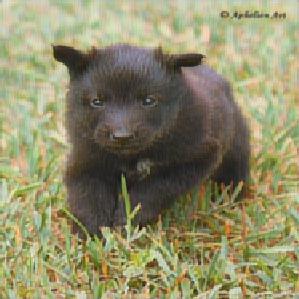} &
\includegraphics[width=\turnheightnew]{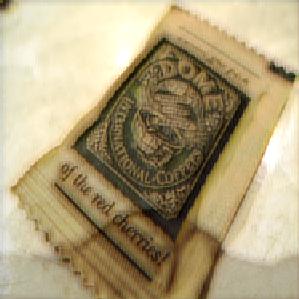} &
\includegraphics[width=\turnheightnew]{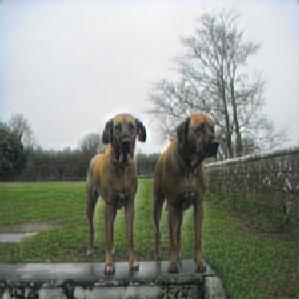} &
\includegraphics[width=\turnheightnew]{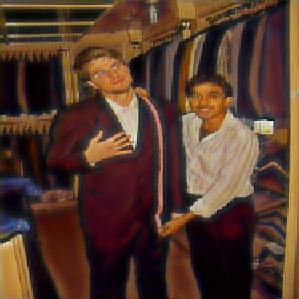}\\
\imlabel{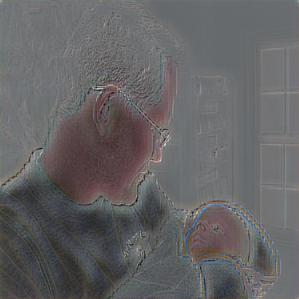}{\textcolor{white}{$\boldsymbol{x}^{\prime} - \boldsymbol{x}$}} &
\includegraphics[width=\turnheightnew]{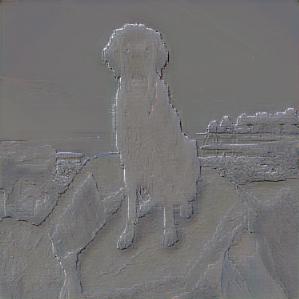} &
\includegraphics[width=\turnheightnew]{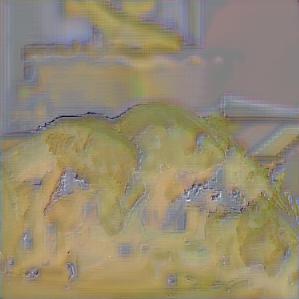} &
\includegraphics[width=\turnheightnew]{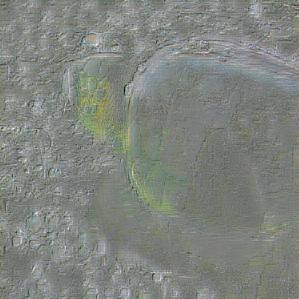} &
\includegraphics[width=\turnheightnew]{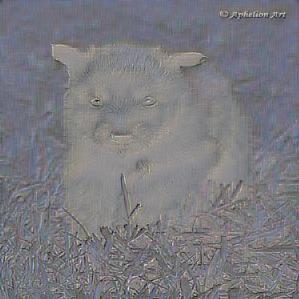} &
\includegraphics[width=\turnheightnew]{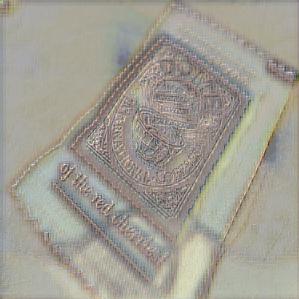} &
\includegraphics[width=\turnheightnew]{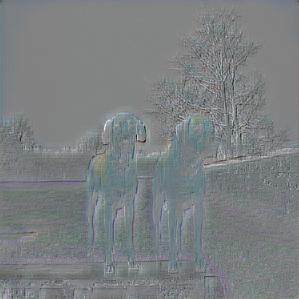} &
\includegraphics[width=\turnheightnew]{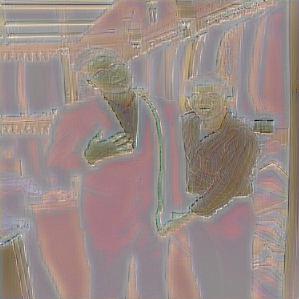}\\
\end{tabular}

  \caption{\textbf{Adversarial examples of GA-DMI-FSA}, which are misclassified by \emph{all} the models in Table.~\ref{benchmark}. Top: benign examples $\boldsymbol{x}$. Middle: unrestricted adversarial examples $\boldsymbol{x}^{\prime}$. Bottom: normalized adversarial perturbations $\boldsymbol{x}^{\prime} - \boldsymbol{x}$.}
  \label{fig:semantic}
  \vspace{-4pt}
\end{figure*}

\begin{itemize}
    \item We propose a Geometry-Aware (GA) framework, where fixed-budget attacking methods can be integrated, to generate transferable {unrestricted} adversarial examples with {approximately} minimum changes. To the best of our knowledge, we are the first to explore transfer-based black-box attacks with adaptive perturbation budgets. 
    
    \item Under $\ell_{\infty}$-norm setting, our GA framework improves the imperceptibility of the crafted adversarial examples by a large margin without the decrease of transfer success rate (see Fig.~\ref{fig:score}). By applying our method GA-DTMI-FGSM to the \emph{CVPR’21 Security AI Challenger}~\cite{chen2021unrestricted}, we ranked 1st place out of 1,559 teams and surpassed the runner-up submissions by 4.59\% and 23.91\% in terms of final score and average image quality level, respectively.
    
    \item Under unrestricted setting, we propose a transfer-based unrestricted attack (see Eq.~\eqref{unrestricted}) by combining the \emph{white-box} feature space attack~\cite{Xu_Tao_Cheng_Zhang_2021} with transfer-based $\ell_{\infty}$-norm attacks to generate semantic-preserving {yet} transferable adversarial examples (see Figs.~\ref{fig:insight} and \ref{fig:semantic}). Moreover, the crafted adversarial examples transfer well to adversarially robust models (see Table.~\ref{benchmark}, Figs.~\ref{fig:attack_adv} and \ref{fig:adv_more}).
\end{itemize}

\section{Related Work}
\label{sec:related}

\subsection{$\ell_p$-norm Adversarial Examples}
\label{sec:attack} 
Existing gradient-based white-box attacks either search for adversarial examples in a fixed $\ell_p$-norm ball~\cite{madry2017towards,Kurakin2019}, or optimize the perturbation for each image independently to get a \emph{minimum-norm} solution such as DeepFool~\cite{moosavi2016deepfool}, CW~\cite{carlini2017towards}, and fast adaptive boundary attack~\cite{croce2020minimally}. However, white-box assumption usually does not hold in real-world scenarios. In query-based black-box 
setting, attackers utilize output logits~\cite{chen2017zoo,andriushchenko2020square} or predicted label~\cite{brendel2018decisionbased,cheng2018queryefficient} of the target model to generate adversarial examples. But these attacks typically suffer from high query complexity, making it easy to be detected~\cite{willmott2021you}. Transfer-based black-box attacks~\cite{Lin2020Nesterov,Wu2020Skip,wang2021admix,wang2021a} can pose serious threats in practice as they need no information about the defense models. Dong \etal~\cite{dong2018boosting} boosted transferability by integrating momentum into gradient-based methods. Liu \etal~\cite{liu2017delving} found that attacking a group of substituted source models simultaneously can improve transferability. Besides, transferability benefits from input transformations such as input diversity~\cite{xie2019improving} and translation-invariant method~\cite{dong2019evading}. 

\vspace{-2pt}
\subsection{Unrestricted Adversarial Examples}
\label{sec: beyond}
The $\ell_p$-norm distance is not an ideal perceptual similarity metric~\cite{johnson2016perceptual,isola2017image}, which oversimplifies the diversity of real-world perturbations. Unrestricted adversarial examples have received significant attention in recent years~\cite{brown2018unrestricted}. Most of the current unrestricted attacks aim to generate imperceptible adversarial examples under \emph{white-box} setting, such as geometric transformations~\cite{xiao2018spatially,alaifari2018adef,engstrom2019exploring} and distance metrics beyond $\ell_p$ norm~\cite{wong2019wasserstein,laidlaw2021perceptual}. Color-based attacks~\cite{hosseini2018semantic,laidlaw2019functional,zhao2020towards,conf/bmvc/ZhaoLL20,shamsabadi2020colorfool,Bhattad2020Unrestricted} were also proposed to generate large but imperceptible perturbations, however, the modified color can sometimes be unnatural. Instead of optimizing in the input space, generative approaches~\cite{NEURIPS2018_8cea559c,gowal2020achieving,qiu2020semanticadv,wong2020learning} search for adversarial embeddings in the latent space. Style transfer~\cite{prabhu2018art,Bhattad2020Unrestricted} is inherently an unrestricted attack as it preserves the semantic of the content image. However, constructing transferable unrestricted adversarial examples is still less explored.  In this work, we will fill this gap by combining the white-box feature space attack~\cite{Xu_Tao_Cheng_Zhang_2021} with transfer-based $\ell_{\infty}$-norm attacks to generate semantic-preserving {yet} transferable unrestricted adversarial examples.

\vspace{-2pt}
\subsection{Adversarial Defenses}
\label{sec:defense}
There have been long-standing arms races between defenders and attackers.
\emph{Adversarial training}~\cite{DBLP:journals/corr/GoodfellowSS14} is one of the most promising defense methods. Many variants of adversarial training framework were proposed, e.g., ensemble adversarial training~\cite{liu2017delving} for transfer-based attacks, PGD-based adversarial training~\cite{madry2017towards}, and TRADES~\cite{zhang2019theoretically} with a new robust loss based on the trade-off between robustness and accuracy. Geometry-aware instance-reweighted adversarial training~\cite{zhang2021_GAIRAT}, which is proven falling into gradient masking~\cite{hitaj2021evaluating}, shares similar insights with us that the importance of distinct inputs in adversarial training should be different. Laidlaw \etal~\cite{laidlaw2021perceptual} integrate adversarial training with Learned Perceptual Image Patch Similarity (LPIPS)~\cite{zhang2018unreasonable}, aiming to improve robustness against perturbations that were unseen during training. Unlike empirical defenses,  {Certified defenses}~\cite{cohen2019certified,10.5555/3454287.3455300,zhang2020towards,leino21gloro} could provide robustness guarantee under a certain $\ell_p$-norm budget. 


\section{Preliminaries} 
\label{sec:Preliminaries}

\textbf{Notation.}
A deep neural network classifier can be described as a function $f(\boldsymbol{x};\boldsymbol{\theta}): \mathcal{X} \rightarrow \mathbb{R}^{\mathcal{C}}$, parameterized by weights $\boldsymbol{\theta}$, which maps a vector $\boldsymbol{x} \in \mathcal{X}$ to its output logits. Given an input $\boldsymbol{x}$ of class $y \in \{1,2,\cdots, \mathcal{C}\}$, the predicted label of $f(\boldsymbol{x};\boldsymbol{\theta})$ is $\hat{f}(\boldsymbol{x}) := \arg \max_j f_j(\boldsymbol{x};\boldsymbol{\theta})$, where $f_j(\boldsymbol{x};\boldsymbol{\theta})$ represents the $j$-th entry of $f(\boldsymbol{x};\boldsymbol{\theta})$. We use $L\left(f(\boldsymbol{x};\boldsymbol{\theta}), y\right)$ to represent the cross-entropy loss and denote the $\varepsilon$-neighborhood of $\boldsymbol{x}$ by $\mathbb{B}(\boldsymbol{x}, \varepsilon):= \{\boldsymbol{x}^{\prime} \in \mathcal{X}: \mathcal{D}(\boldsymbol{x}, \boldsymbol{x}^{\prime}) \le \varepsilon\}$, where $\mathcal{D}$ is a distance metric that describes the change between the adversarial example $\boldsymbol{x}^{\prime}$ and the nature example $\boldsymbol{x}$. We denote the black-box test model by $g$, and split the set of accessible source models $\Phi = \{\phi_1, \phi_2, \cdots, \phi_n\}$ into the set of training models $f$ and the set of validation models $h$.


\subsection{Transfer-based $\ell_{\infty}$-norm Attacks}
\label{sec:linf-transfer} 
Existing transfer-based attacks typically search for adversarial examples in a fixed-radius $\ell_{p}$-norm ball, i.e., $\mathcal{D}(\boldsymbol{x}, \boldsymbol{x}^{\prime}) = \|\boldsymbol{x}^{\prime} - \boldsymbol{x}\|_p \le \varepsilon$. Various methods were proposed to boost transferability of the generated adversarial examples, such as input Diversity Iterative Fast Gradient Sign Method (DI-FGSM)~\cite{xie2019improving}, Momentum-based Iterative (MI-FGSM) method~\cite{dong2018boosting} and Translation-invariant Iterative (TI-FGSM) method~\cite{dong2019evading}. We formulate a strong $\ell_{\infty}$-norm baseline DTMI-FGSM by combining all these techniques, i.e., 
\begin{equation} \label{DTMI}
\begin{aligned}
    \boldsymbol{m}_{t+1} & = \gamma \cdot \boldsymbol{m}_t + \frac{\mathbf{W} * \nabla_{\boldsymbol{x}_t}L\left(f\left(T(\boldsymbol{x}_t, p);\boldsymbol{\theta}\right), y\right)}{\|\mathbf{W} * \nabla_{\boldsymbol{x}_t}L\left(f\left(T(\boldsymbol{x}_t, p);\boldsymbol{\theta}\right), y\right)\|_1}, \\
    \boldsymbol{x}_{t+1} & = \Pi_{\mathbb{B}(\boldsymbol{x}, \varepsilon)}\left(\boldsymbol{x}_{t} + \alpha \cdot \texttt{sign}(\boldsymbol{m}_{t+1})\right),
\end{aligned}
\end{equation}
where $\boldsymbol{m}_0 = \mathbf{0}$, $\mathbf{W}$ is a pre-defined kernel with a convolution operation $*$, $\alpha$ is the step size, $\Pi$ is the projection operator, and $\gamma$ is the decay factor for the momentum term. $T(\boldsymbol{x}_t, p)$ represents the input transformation on $\boldsymbol{x}_t$ with probability $p$. When $\gamma = 0$, DTMI-FGSM attack degenerates to the DTI-FGSM attack. When $p = 0$, DTMI-FGSM attack degenerates to the DMI-FGSM attack. 

\subsection{Transfer-based Unrestricted Attack}
\label{sec:beyonglp}

Inspired by prior work~\cite{huang2017arbitrary} in style transfer, Xu \etal~\cite{Xu_Tao_Cheng_Zhang_2021} tries to find stylized adversarial examples by assuming that the image pairs from the same class share consistent content and differ mainly in their styles. Here we propose to generate semantic-preserving {yet} transferable unrestricted adversarial examples by combining the Feature Space Attack (FSA)~\cite{Xu_Tao_Cheng_Zhang_2021} with transfer-based $\ell_{\infty}$-norm attacks~\cite{dong2018boosting,xie2019improving}. Given an encoder $\phi$, we extract the style features of input $\boldsymbol{x}$ as channel-wise mean $\boldsymbol{\mu}\left(\phi(\boldsymbol{x})\right) \in \mathbb{R}^C$ and channel-wise standard deviation $\boldsymbol{\sigma}\left(\phi(\boldsymbol{x})\right) \in \mathbb{R}^C$, 
Specifically, 
\begin{equation}
\begin{aligned}
    \mu_c & =  \frac{1}{HW}\sum_{h=1}^H\sum_{w=1}^W \phi_c(\boldsymbol{x})_{hw}, \\
     \sigma_c & =  \sqrt{\frac{1}{HW}\sum_{h=1}^H\sum_{w=1}^W \left(\phi_c(\boldsymbol{x})_{hw} - \mu_c\right)^2},
\end{aligned}
\end{equation}
where $\phi(\boldsymbol{x}) \in \mathbb{R}^{C\times H \times W}$ represents the latent embedding. Xu \etal~\cite{Xu_Tao_Cheng_Zhang_2021} adds adversarial perturbations on $\boldsymbol{\mu}$ and $\boldsymbol{\sigma}$ before projecting  $\phi(\boldsymbol{x})$ back to the input space $\mathcal{X}$ with a pre-trained\footnote{We use the official pre-trained shallowest decoder: \url{https://github.com/qiulingxu/FeatureSpaceAttack}.} decoder $\phi^{-1}$, namely, 
\begin{equation}
\begin{aligned}
 \tilde{\phi}(\boldsymbol{x}) & =  
e^{\boldsymbol{\tau}^{\sigma}} \cdot  \left(\phi(\boldsymbol{x}) - \boldsymbol{\mu} \right)+ e^{\boldsymbol{\tau}^{\mu}} \cdot  \boldsymbol{\mu},\\
\boldsymbol{x}^{\prime} & = \phi^{-1} \left(\tilde{\phi}(\boldsymbol{x}) \right), \quad \|\boldsymbol{\tau}^{\mu}\|_{\infty} \le \ln \varepsilon, \|\boldsymbol{\tau}^{\sigma}\|_{\infty} \le \ln \varepsilon,
\end{aligned}
\label{unrestricted}
\end{equation}
where $ \tilde{\phi}(\boldsymbol{x})$ enlarges or shrinks the mean $\boldsymbol{\mu}$ and the standard deviation $\boldsymbol{\sigma}$ of the embedding $\phi(\boldsymbol{x})$ by a factor of $e^{\boldsymbol{\tau}^{\mu}}$ and $e^{\boldsymbol{\tau}^{\sigma}}$, respectively. In this way, the distance metric $\mathcal{D}(\boldsymbol{x}, \boldsymbol{x}^{\prime}) = \max \left(e^{\|\boldsymbol{\tau}^{\mu}\|_{\infty}}, e^{\|\boldsymbol{\tau}^{\sigma}\|_{\infty}}\right) \le \varepsilon$. In order to preserve the semantic of the unrestricted adversarial example $\boldsymbol{x}^{\prime}$, a content loss was added during the attacking process, i.e.,
\begin{equation*}
    \min_{\boldsymbol{\tau}^{\mu}, \boldsymbol{\tau}^{\sigma}} \; \mathcal{L}(\boldsymbol{x}^{\prime}, y)  =  \lambda \cdot \mathcal{L}_{\text{top-5}}\left(f(\boldsymbol{x}^{\prime};\boldsymbol{\theta}), y\right) +  \| \phi(\boldsymbol{x}^{\prime}) - \tilde{\phi}(\boldsymbol{x})\|_2,\notag 
\end{equation*}
where $\lambda$ balance the trade-off between adversarial and the content loss. Following Xu \etal~\cite{Xu_Tao_Cheng_Zhang_2021}, we set $\lambda = 128$ and use the top-5 margin loss for adversarial attack.  
With all above, the unrestricted attack (see Eq.~\eqref{unrestricted}) can be solved by conventional $\ell_{\infty}$-norm attack on parameters $\boldsymbol{\tau}^{\mu}$ and $\boldsymbol{\tau}^{\sigma}$. Moreover, the same techniques in Sec.~\ref{sec:linf-transfer} such as input diversity $T(\boldsymbol{x}^{\prime}, p)$ (only for the margin loss) and momentum-based method can be integrated to improve transferability, i.e.,
\begin{equation} \label{DMI-FSA}
\begin{aligned}
    \boldsymbol{m}_{t+1} & = \gamma \cdot \boldsymbol{m}_t + \frac{ \nabla_{\boldsymbol{\tau}_t}\mathcal{L}  \left( T(\boldsymbol{x}^{\prime}, p), y\right)}{\|\nabla_{\boldsymbol{\tau}_t}\mathcal{L}  \left( T(\boldsymbol{x}^{\prime}, p), y\right)\|_1}, \\
    \boldsymbol{\tau}_{t+1} & = \Pi_{\mathbb{B}(\boldsymbol{x}, \varepsilon)}\left(\boldsymbol{\tau}_{t} - \alpha \cdot \texttt{sign}(\boldsymbol{m}_{t+1})\right).
\end{aligned}
\end{equation}
When $\gamma = 0$, the DMI-FSA attack degenerates to the DI-FSA attack.

\begin{figure*}[t]
  \begin{center}
  \includegraphics[width=1.0\textwidth]{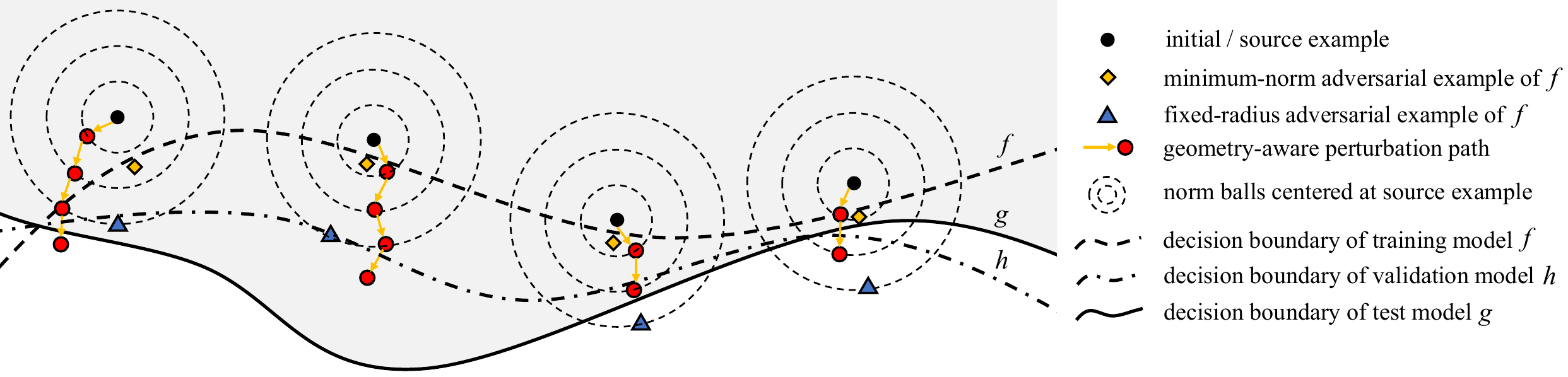}
  \end{center}
  \vspace{-10pt}
  \caption{\textbf{Our geometry-aware framework}. Existing fixed-budget methods typically overlook the importance of geometrical distances from inputs to the decision boundary of the test model $g$. In contrast, our geometry-aware framework aims to find geometry-aware minimal-change perturbation via a validation model $h$. The goal of the validation model is to prevent an attack algorithm overfitting $f$ by forcing the solution to cross the decision boundary of $h$ with a certain margin. Our framework consists of multiple sub-procedures with adaptive perturbation budgets. In each sub-procedure, we start from the solution of the last sub-procedure (the red solid points) and re-run the attack algorithm on the training model $f$. The procedure stops if the output probability of the true class on the validation model $h$ is smaller than a certain threshold $\eta$.}
  \label{fig:conceptual}
\end{figure*}

\subsection{Evaluation Metric for Transfer-based Attack}
\label{sec:transfer}


The imperceptibility of adversarial examples is hard to evaluate due to the lack of precise quantization of human perception~\cite{wang2004image}. Sharif \etal~\cite{sharif2018suitability} found that $\ell_p$-norm distance is not an ideal perceptual similarity metric and suggest setting {adaptive} perturbation budget for every sample to ensure that the attacks’ output would be imperceptible. Therefore, we choose transfer success rate and the perturbation budget under distance metric $\mathcal{D}$ as our main evaluation metrics.
Consider a dataset $\hat{S} = \{(\boldsymbol{x_i}, y_i)\}_{i=1}^N$ and the corresponding adversarial examples $\hat{S}_{\text{adv}} = \{(\boldsymbol{x_i^{\prime}}, y_i)\}_{i=1}^N$ that are crafted on the training model $f$. Let $N_0 = \sum_{i=1}^N \mathbbm{1}\{\hat{g}(\boldsymbol{x}_i^{\prime}) \neq y_i\}$ be the number of misclassified adversarial examples on the test model $g$. We define the average total score as:
\begin{equation} \label{total}
\begin{aligned}
       S_{total} & = \frac{1}{N}\sum_{i=1}^N \mathbbm{1}\{\hat{g}(\boldsymbol{x}_i^{\prime}) \neq y_i\} \cdot \mathcal{F}_{\text{reward}}\left(\mathcal{D}(\boldsymbol{x}_i, \boldsymbol{x}_i^{\prime})\right) \\
       & = \frac{N_0}{N} \cdot \frac{1}{N_0} \sum_{i=1}^N \mathbbm{1}\{\hat{g}(\boldsymbol{x}_i^{\prime}) \neq y_i\} \cdot \mathcal{F}_{\text{reward}}\left(\mathcal{D}(\boldsymbol{x}_i, \boldsymbol{x}_i^{\prime})\right) \\
       & \overset{\underset{\text{def}}{}}{=} \frac{N_0}{N} \cdot  S_{APR},
\end{aligned}
\end{equation}
where $S_{APR}$ is the Average Perturbation Reward of adversarial examples that are misclassified by test model $g$ and the reward function $\mathcal{F}_{\text{reward}}$ is a decreasing function w.r.t. metric $\mathcal{D}(\boldsymbol{x}, \boldsymbol{x}^{\prime})$.

\section{Methodology: Geometry-Aware Framework}
\label{sec:Methodology}

\label{sec:lp}

\begin{algorithm}[h] 
\caption{Geometry-Aware Framework}
\label{alg:Framwork} 
\begin{algorithmic}[1]
\REQUIRE ~~\\
	Benign input $\boldsymbol{x}$ with label $y$; training models $f$; validation model $h$; number of sub-procedures $K$; maximum perturbation size $\varepsilon$ and threshold $\eta$; attack algorithm $\mathcal{A}$;
\ENSURE ~~\\
	Transfer-based unrestricted adversarial example $\boldsymbol{x}^{\prime}$ with {approximately} minimum change;
	\STATE $\boldsymbol{x}_0 = \boldsymbol{x}$;\
	\FOR{$k = 1,2,\cdots, K$}
	    \STATE $\boldsymbol{x}_k = \mathcal{A}(\boldsymbol{x}, \boldsymbol{x}_{k-1},f, \frac{k\varepsilon}{K})$; {\hfill\texttt{$\triangleright$ fixed budget}} \\
	    \STATE $\operatorname{conf} \leftarrow \frac{\exp\left(h_y(\boldsymbol{x}_k;\boldsymbol{\theta})\right)}{\sum_{j} \exp\left(h_j(\boldsymbol{x}_k ;\boldsymbol{\theta})\right)} $;\
	    \IF{$\operatorname{conf} < \eta$}
	        \STATE $\boldsymbol{\text{Return}}$ $\boldsymbol{x}_k$; {\hfill \texttt{$\triangleright$ early-stopping in Eq.~\eqref{early}}}
	    \ENDIF
	\ENDFOR
\STATE $\boldsymbol{\text{Return}}$ $\boldsymbol{x}_K$;
\end{algorithmic}
\end{algorithm}


Eq.~\eqref{total} factorizes the average total score as the product of transfer success rate and average perturbation reward, which motivates us to find the adversarial example with minimum changes under metric $\mathcal{D}$, i.e.,
\begin{equation}
    \begin{aligned}
        \min_{\boldsymbol{x}^{\prime}} \; \mathcal{D}(\boldsymbol{x}, \boldsymbol{x}^{\prime}), \qquad 
        \text{s.t.}  \;\; \hat{g}(\boldsymbol{x}^{\prime}) \neq y.
    \end{aligned}
    \label{eq: argmin}
\end{equation}
However, direct optimization of problem~\eqref{eq: argmin} is intractable, in part due to the lack of information about test model $g$. We approximately solve this problem by discretizing the continuous space of perturbation radius into a discrete set and choosing the minimum perturbation budget such that the attack is able to fool the test model $g$. However, the challenge is that it is typically difficult to decide whether a given perturbation radius can also fool the test model~\cite{NEURIPS2019_32508f53,katzir2021s}. This problem is also known as model selection (we view the source models as training data, then the generated adversarial perturbation is the so-called \textit{selected model} or optimized parameters), and a classic approach to tackle this problem is to have a validation set. More specifically, we split all accessible source models into training model set and validation model set. With validation model $h$, we are able to generate transferable adversarial examples with dynamic radii. To approximately solve problem~\eqref{eq: argmin}, we first divide the attack in the ball $\mathbb{B}(\boldsymbol{x}, \varepsilon)$ into $K$ sub-procedures. In the $k$-th sub-procedure, we re-run a fixed-radius attack algorithm $\mathcal{A}$ such as DMI-FSA (see Eq.~\eqref{DMI-FSA}) under the perturbation budget
\begin{equation*}
    \varepsilon_k = \frac{k}{K}\times \varepsilon,\ k = 1,2,\cdots, K.
\end{equation*}
Each sub-procedure starts from the solution of last sub-procedure to accelerate the convergence. To obtain a minimum-radius solution, we perform an early-stopping mechanism at the end of each sub-procedure if the probability of true class on the validation model $h$ is smaller than a threshold $\eta$, i.e., 
\begin{equation} \label{early}
\mathbb{P} \left(\hat{h}(\boldsymbol{x}) = y\right) =     \frac{\exp\left(h_y(\boldsymbol{x};\boldsymbol{\theta})\right)}{\sum_{j} \exp\left(h_j(\boldsymbol{x};\boldsymbol{\theta})\right)} < \eta.
\end{equation}

\begin{table*}[t]
\small
  \centering
  \caption{\textbf{An overview of all considered networks for generating adversarial examples}. $\text{Top-1}_{\text{ImageNet}}$ represents the accuracy on the ILSVRC 2012 validation set while $\text{Top-1}_{\text{1000}}$ represents the accuracy on the randomly selected 1000 images.}
  \resizebox{1.0\textwidth}{!}{
    \begin{tabular}{c||cccc||cccc}
    \hline
    Training  &  Index & Model Name &  $\text{Top-1}_{\text{ImageNet}}$ &  $\text{Top-1}_{\text{1000}}$ & Index & Model Name  &  $\text{Top-1}_{\text{ImageNet}}$ &  $\text{Top-1}_{\text{1000}}$  \\
    \hline
    \hline
    \multirow{4}[1]{*}{Normal} & 0     &   ViT-S/16  &  76.01\%  &  99.8\%  & 1     &   ViT-B/16 & 81.08\% & 99.1\% \\
          & 2     &   Swin-B/patch4-window7 &  \textbf{84.23\%} &  99.4\% & 3     & ResNeXt101-32x8d-swsl & 83.62\% & 99.9\% \\
          & 4     &   ResNeXt50-32x4d-ssl &  78.90\%  &  99.7\%  & 5     & ResNet50-swsl & 79.97\% & 99.5\% \\
          & 6     &    Inception-v3  & 76.94\% & 100.0\% & 7     & Inception-ResNet-v2 & 79.85\% & 99.9\%\\
    \hline
    \hline
    Ensemble & 8     &    Ens3-adv-Inception-v3  & 76.49\%  & 100.0 \%& 9     & Ens-adv-Inception-ResNet-v2 & 78.98\% & 99.9\% \\
    \hline
    \end{tabular}%
    }
  \label{tab:models}%
\end{table*}%

Our GA framework is summarized in Algorithm~\ref{alg:Framwork} and illustrated in Fig.~\ref{fig:conceptual}. Note that the output of GA framework is related to the choice of the training model $f$ and the validation model $h$. Thus it is important to figure out \textit{which partition of the source models performs better}.

\begin{figure}[t]
  \begin{center}
  \includegraphics[width=0.9\columnwidth]{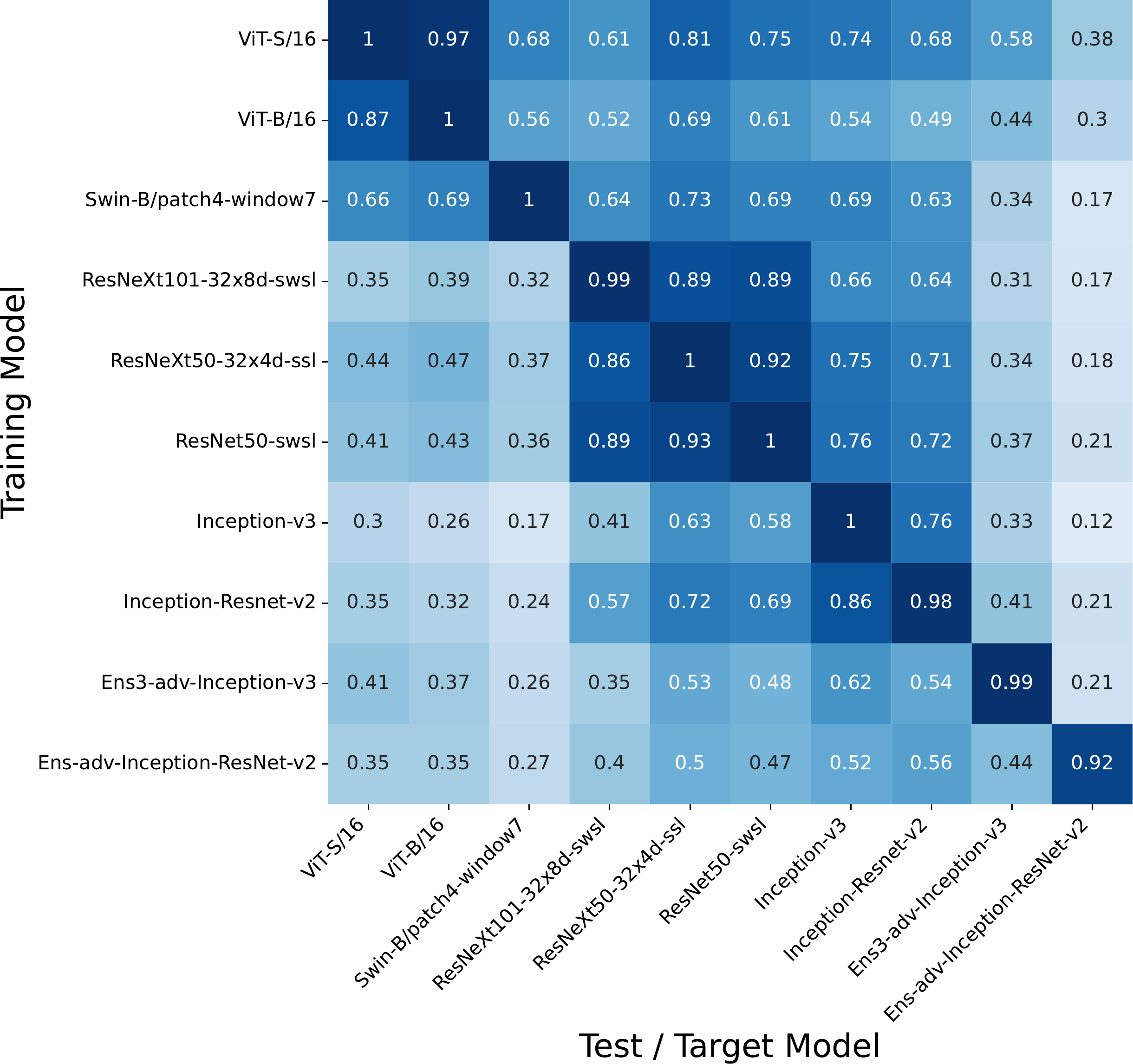}
  \end{center}
  \vspace{-10pt}
  \caption{\textbf{Transferability between networks under DTMI-FGSM attack}. The rows stand for source models and the columns stand for target models. Adversarial examples transfer well between models with similar architectures.}
  \vspace{-10pt}
  \label{fig:attack_models}
\end{figure}

We split $n$ pre-trained models $\Phi = \{\phi_1, \phi_2, \cdots, \phi_n\}$ into $k$ training models and $n - k$ validation models. Instead of traversing all possible partitions to select the optimal split by querying the test model $g$ by $C_n^k$ times, we propose a query-free approach that only 
utilizes the information of transferability between the pre-trained models (see Fig.\ref{fig:attack_models}). Let $w_{ij}$ ($w_{ij} \neq w_{ji}$) be the transfer success rate ($\frac{N_0}{N}$ in Eq.~\eqref{total}) from the source model $\phi_i$ to the target model $\phi_j$ under a fixed-radius attack (e.g., DTMI-FGSM). Denote the binary partition function as $\mathcal{G}$. The training set and the validation set can be formulated as $\mathcal{T} = \{i \mid \mathcal{G}(\phi_i) = 0\}$ and $\mathcal{V} = \{j \mid \mathcal{G}(\phi_j) = 1\}$, respectively. We define the \emph{partition loss} $\ell_{\mathcal{G}}$ as:
\begin{equation}
\begin{aligned}
\ell_{\mathcal{G}} & = \frac{1}{k}\sum_{i \in \mathcal{T}}\ell_{\mathcal{G}}(\phi_i)  + \frac{1}{n - k}\sum_{j \in \mathcal{V}}\ell_{\mathcal{G}}(\phi_j), \\
\ell_{\mathcal{G}}(\phi_i) & = \frac{1}{ k - 1}\sum_{t \neq i, t \in \mathcal{T}} w_{it} + \frac{1}{n - k}\sum_{t \in \mathcal{V}}  w_{it},\\
\ell_{\mathcal{G}}(\phi_j) & = \frac{1}{ n-k - 1}\sum_{t \neq j, t \in \mathcal{V}} w_{jt} + \frac{1}{k}\sum_{t \in \mathcal{T}} w_{tj}.
\end{aligned}
\label{diversity_formula}
\end{equation}
For both the training set loss $\ell_{\mathcal{G}}(\phi_i)$ and the validation set loss $\ell_{\mathcal{G}}(\phi_j)$: 1) Minimizing the first formula on the right of Eq.~\eqref{diversity_formula} encourages intra-group diversity. To make the decision boundary of the ensemble model ($f$ or $h$) more general and effective, we minimize the transfer success rate between any two pre-trained models inside the group. 2) Minimizing the second formula on the right of Eq.~\eqref{diversity_formula} is penalizing extra-group similarity. If $f = h$, the early-stopping mechanism in Eq.~\eqref{early} will be triggered too early, leading to small adversarial perturbations for all inputs. Reducing the transferability from the training model $f$ to the validation model $h$ might improve generalization of the adversarial examples to unknown models. Empirically, we find that the proposed partition loss $\ell_{\mathcal{G}}$ negatively correlates with the average total score $S_{total}$.



\section{Experiments} \label{Exp}

\subsection{Experimental Setup}
\label{sec:experiments setup}

\medskip
\noindent\textbf{Datasets \& Networks.} Similar to Xie \etal~\cite{xie2019improving}, we randomly select 1,000 images from ILSVRC 2012 validation set \cite{deng2009imagenet}, which are almost correctly classified by all the attacking models. All these images are resized to $229 \times 229 \times 3$ beforehand. We consider eight normally trained models and two ensemble adversarially trained models~\cite{tram2018}. The weights of all these models are publicly available~\cite{rw2019timm}. More details about the networks are summarised in Table.~\ref{tab:models}. The transferability between these models under $\ell_{\infty}$-norm setting is summarised in Fig.~\ref{fig:attack_models}. It is much easier for the generated adversarial examples to transfer from vision transformers to convolutional neural networks (CNNs), which is consistent with the empirical observation in Shao \etal~\cite{shao2021adversarial}. Surprisingly, the robustness of naturally trained vision transformers under transfer attack is even on par with two ensemble adversarially trained CNNs.


\medskip
\noindent\textbf{Implementation Details.} Given the maximum perturbation size $\varepsilon$ and number of sub-procedures $K$ (5 as default) in our geometry-aware framework, we set the step size $\alpha = \frac{1.25  \times \varepsilon_k}{T}$ in the $k$-th sub-procedure, where the number of iterations $T$ is set to 10 in the $\ell_{\infty}$ setting and 50 in the unrestricted setting. $\varepsilon$ is set to 20 in the $\ell_{\infty}$-norm setting and 3.5 in the unrestricted setting. When running a fixed-radius baseline at perturbation budget $\varepsilon_k$, we set the number of iteration as $\frac{T}{2} \left(1+ \frac{K \varepsilon_k}{\varepsilon}\right)$ with step size $\alpha$ to keep the same total perturbation budget (the sum of step size across all iterations) as our geometry-aware framework for fair comparison. The reward function $\mathcal{F}_{\text{reward}}(\varepsilon_0)$ is set to  $1 / \varepsilon_0$ as smaller perturbation radius exhibits significantly higher image quality. For the momentum term, we set the decay factor $\mu = 1$ as in Dong \etal~\cite{dong2018boosting}. For DI-FGSM~\cite{xie2019improving}, we set the transformation probability to $p = 0.7$. The input is first randomly resized to be an $r \times r \times 3$ image with $r \in [(1 - \gamma)s, (1 + \gamma)s]$, and then padded to size $(1 + \gamma)s \times (1 + \gamma)s \times 3$. The transformed input is then resized to $s \times s \times 3$ for different input size $s$ of various models, i.e., 224, 299 and 384. We set $\gamma = 0.1$ as default. For TI-FGSM~\cite{dong2019evading}, we use Gaussian kernel with kernel size $5 \times 5$.

\subsection{{Balancing} Transfer Success Rate and Perturbation Reward}
\label{balance}

\begin{figure}[t]
  \begin{center}
  \includegraphics[width=0.90\columnwidth]{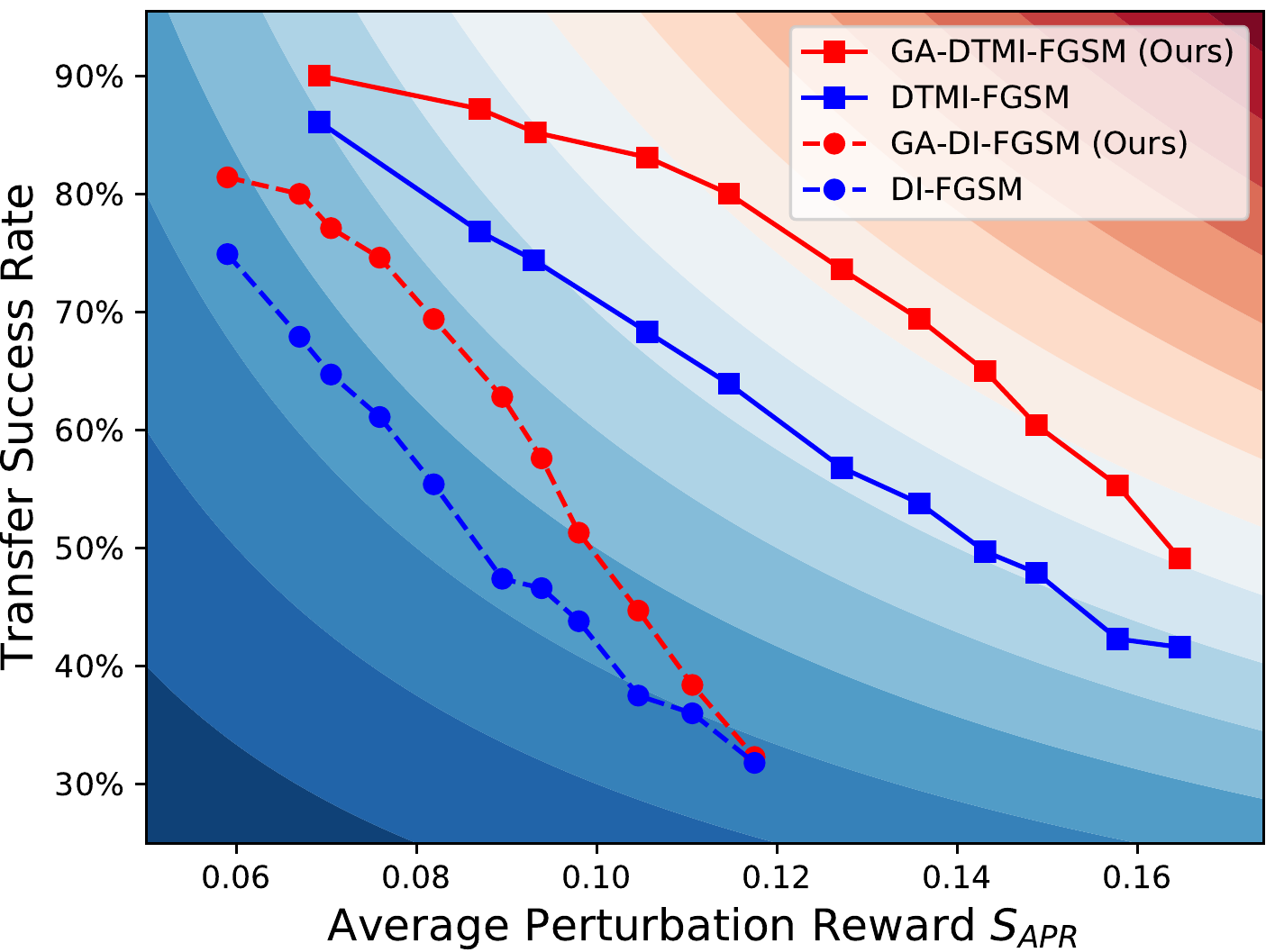}
  \end{center}
  \vspace{-10pt}
  \caption{\textbf{Contour of average total score $S_{total}$ (higher is better)}. Fixing transfer success rate as 80\%, our approach GA-DTMI-FGSM surpasses the baseline DTMI-FGSM (see Eq.~\eqref{DTMI}) by up to 43.35\% in terms of average perturbation reward.} 
   \label{fig:score} 
\end{figure}

\medskip
\noindent\textbf{Implementation Details.} Benefiting from the adaptive choice of perturbation budgets, our geometry-aware framework can generate transferable {unrestricted} adversarial examples with smaller changes. In this experiment, the training model $f$ and validation model $h$ are an ensemble of models $\{2,3,5\}$ and $\{1,4,6\}$ in Table.~\ref{tab:models}, respectively. The test model is Inception-ResNet-v2. {The optimal threshold $\eta$ (see Eq \eqref{early}) is searched from a finite set ranging from 0.001 to 0.9 by querying\footnote{In contrast to conventional query-based attacks that need the logits or predicted label on the target model, we query whether an adversarial example transfers to the target model successfully. Besides, we have prior information on $\eta$ which depends on the similarity between $f,h$ and $g$, making the query complexity rather limited.} the test model $g$ to achieve the best average total score $S_{total}$. For each $\eta$, we execute our method and compute the average perturbation reward $S_{APR}$ (the $x$-axis of each red point in Fig.~\ref{fig:score}). Then the corresponding fixed-radius baseline is run at the same $x$-axis}. We conduct experiments on two threat models. For the $\ell_{\infty}$-norm setting, we combine our Geometry-Aware (GA) framework with DI-FGSM, DTI-FGSM, DTMI-FGSM, and Admix-DTI-FGSM~\cite{wang2021admix} (limited by the memory of a single NVIDIA RTX 3090, we set the number of admixed images $m_1 = 3$ and the number of randomly sampled images from other categories $m_2 = 2$), named GA-DI-FGSM, GA-DTI-FGSM, GA-DTMI-FGSM, and GA-Admix-DTI-FGSM, respectively; For the unrestricted setting, we combine our GA framework with DMI-FSA and DI-FSA, named GA-DMI-FSA and GA-DI-FSA, respectively.

\medskip
\noindent\textbf{Experimental Results.} We present the contour plot of average total score in Fig.~\ref{fig:score}, where the improvement of our method upon baselines depends on the choice of hyper-parameter $\eta$ (leading to different $S_{APR}$). Fixing $S_{APR}$ as 0.115, our approach GA-DTMI-FGSM surpasses the baseline DTMI-FGSM by up to 16.1\% in terms of transfer success rate. As shown in Table.~\ref{tab:tradeoff}, our approach yields a significant performance boost on the average total score $S_{total}$ across various threat models, especially in the $\ell_{\infty}$-norm setting where both the transfer success rate and $S_{APR}$ are improved.

\begin{table}[bh]
\centering
\caption{\textbf{Comparison of our method with baselines}. We report the results when both our approach and baselines achieve highest average total score $S_{total}$. TSR: Transfer Success Rate.}
\vspace{-5pt}
\label{tab:tradeoff} 
\resizebox{0.99\columnwidth}{!}{%
    \begin{tabular}{|c||ccc|}
    \hline
    Method  & TSR ($\uparrow$) & $S_{APR}$ ($\uparrow$)  & $S_{total}$ ($\uparrow$) \\
    \hline
    \hline
  DI-FGSM~\cite{xie2019improving}    & 61.1\%    & 0.0759 & 4.64\% \\
  GA-DI-FGSM  & \textbf{69.4\%}       & \textbf{0.0819} &  \textbf{5.68\%} \\
  \hline
  DTI-FGSM~\cite{dong2019evading}   &  57.3\%  & 0.1101 & 5.68\% \\
  GA-DTI-FGSM  & \textbf{67.9\%}     & \textbf{0.1176} & \textbf{7.98\%} \\
  \hline
  DTMI-FGSM    & 63.9\%    & 0.1147 & 7.33\% \\
  GA-DTMI-FGSM  & \textbf{69.4\%}    & \textbf{0.1358}  & \textbf{9.42\%} \\
\hline
 Admix-DTI-FGSM~\cite{wang2021admix}   &  68.1\% &  0.1248 &  8.50\%\\
  GA-Admix-DTI-FGSM   &  \textbf{82.5\%}  &\textbf{0.1299} & \textbf{10.72\%}\\
    \hline
    \hline
  DI-FSA    & 48.3\%    & 0.5328 & 25.73\% \\
  GA-DI-FSA  & \textbf{50.4\%}    & \textbf{0.5541}  & \textbf{27.93\%} \\
  \hline
DMI-FSA    & 51.3\%    & \textbf{0.5616} & 28.81\% \\
  GA-DMI-FSA   & \textbf{58.3\%}    &  {0.5355} & \textbf{31.32\%} \\
  \hline
    \end{tabular}%
    } 
\end{table}





\subsection{Case Study: CVPR’21 Security AI Challenger} \label{conference}

\begin{table*}[bp]
\small
 \centering
 \vspace{-2pt}
\caption{\textbf{Benchmarking classification on Imagenet under transfer-based unrestricted attacks}. $\text{PGD}_{\text{40}}^*$ indicates the PGD attack with 40 steps ($\varepsilon=\frac{4}{255}$). We denote the adversarial examples crafted by GA-DTMI-FGSM, GA-Admix-DTI-FGSM, and GA-DMI-FSA in Table.~\ref{tab:tradeoff} as $\text{GA}_{\text{DTMI-FGSM}}$, $\text{GA}_{\text{Admix-DTI}}$, and $\text{GA}_{\text{DMI-FSA}}$, respectively. \textbf{Bold} and \underline{underline} indicate the lowest and second lowest in each row.}
\vspace{-2pt}
  \resizebox{0.99\textwidth}{!}{
   \begin{tabular}{l||cc|cc|ccc}
    \hline
    \hline
    Defenses & Clean & $\text{PGD}_{\text{40}}^*$ & ReColor & $\text{FSA}$ & $\text{GA}_{\text{DTMI-FGSM}}$  & $\text{GA}_{\text{Admix-DTI}}$ &  $\text{GA}_{\text{DMI-FSA}}$ \\
    \hline
    \hline 
    $\text{Inception-ResNet-v2}_{\text{Ens-adv}}$~\cite{tram2018}  &   99.9\%          &     \textbf{10.0\% }  & 93.2\% &  90.0\%   &       87.1\%    &  78.1\% &  \underline{43.5\%}  \\
     $\text{Fast}_{\text{AT}}$~\cite{Wong2020Fast}   &    66.9\%    &   \underline{35.7\%}    & 62.8\%&  59.3\%    &  64.7\%     &  64.1\%   &  \textbf{28.5\%}  \\
    $\text{Free}_{\text{AT}}$~\cite{2019arXiv190412843S}   &   77.3\%   &   \underline{40.6\%}     & 71.6\%&   68.4\%     &    74.0\%    &     73.0\%&  \textbf{35.1\%}  \\
    Resnet152-Base~\cite{xie2019feature}  &   67.6\%    &   \underline{39.0\%}   & 64.1\% &  61.2\%    &      65.1\%     &  65.7\% &  \textbf{37.4\%}   \\
    Resnext101-DenoiseAll~\cite{xie2019feature}    & 80.3\%  & \underline{52.2\%}  & 77.0\% & 73.8\% &  78.7\% & 78.1\% & \textbf{47.8\%}  \\
    Resnet152-Denoise~\cite{xie2019feature}   &   72.2\%    &     \underline{41.8\%}      & 68.2\% &  65.2\%    &   70.7\%     & 69.9\% &   \textbf{40.3\%}  \\
    \hline
    RVT-Tiny~\cite{mao2021rethinking}   &    96.7\%    &       \textbf{0.0\%}    & 78.9\% &   81.3\%  &  44.5\%      &    \underline{26.3\%}   & {33.6\%}  \\
    DeepAugment+AugMix~\cite{hendrycks2021many}  &   96.1\%        &  \textbf{0.0\%}       & 82.8\% &  89.3\%     &   {58.9\%}        & \underline{41.4\%} &  63.2\%  \\
    Efficientnet-l2-ns~\cite{xie2020self}  &   99.5\%   &     \textbf{0.0\%}     & 95.7\% &   97.0\%    &     {81.3\%}   &   \underline{78.6\%} & {86.0\%} \\
    Swin-L/patch4-window-12~\cite{liu2021Swin}   &    99.1\%         &     \textbf{0.0\%}    &  88.4\% &  90.7\%    &      66.6\%  & \underline{58.7\%}  &  {61.8\%}  \\
    \hline
    \hline
    \end{tabular}%
    }
  \label{benchmark}%
\end{table*}%

\begin{figure}[t]
  \begin{center}
  \includegraphics[width=0.95\columnwidth]{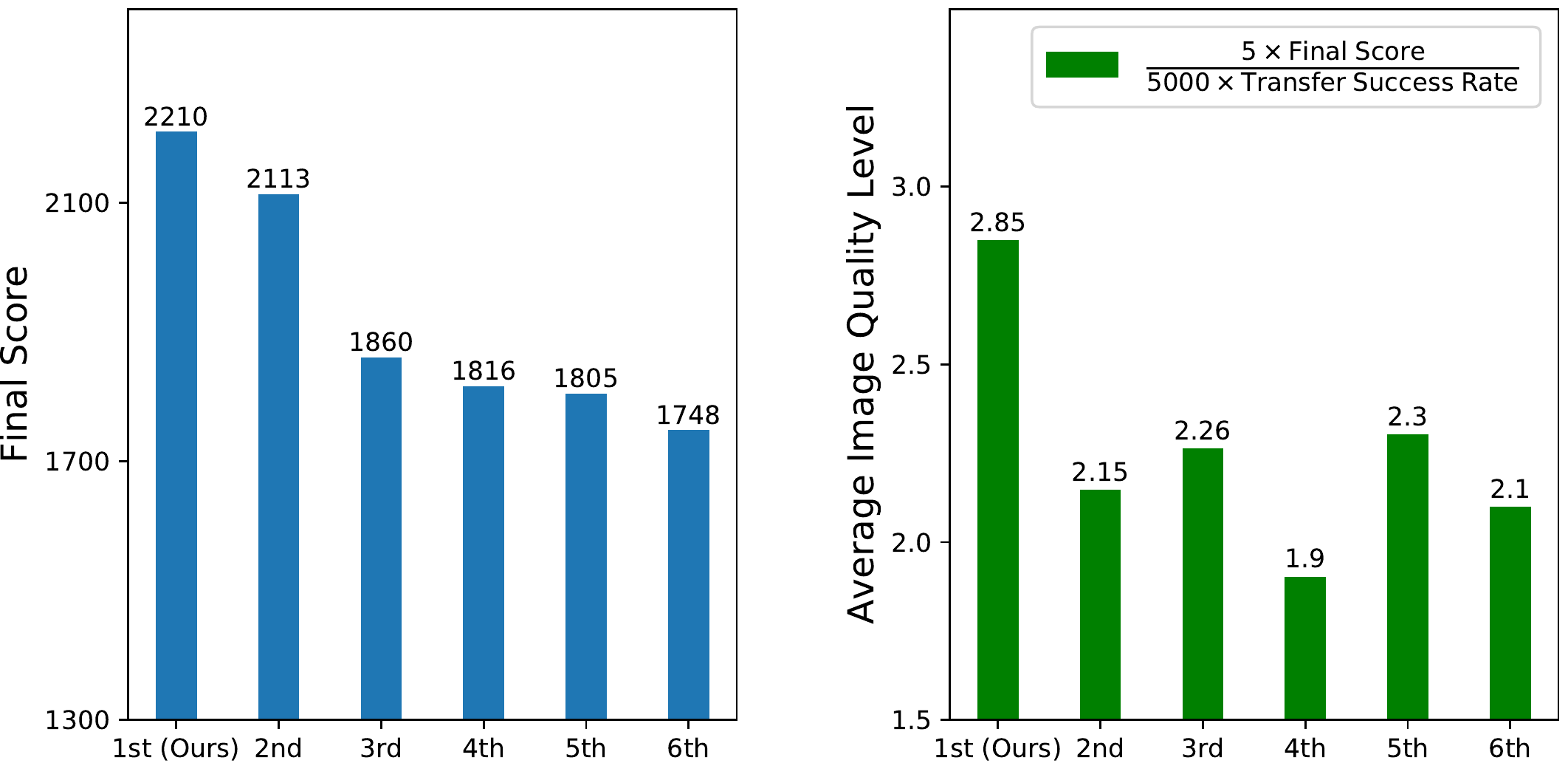}
  \end{center}
  \vspace{-10pt}
  \caption{\textbf{Top-6 results} in the CVPR'21 Security AI Challenger: Unrestricted Adversarial Attacks on ImageNet. The final scores were manually scored by multiple human referees.}
  \label{fig:cvpr_1st}
  \vspace{-5pt}
\end{figure}

In the \emph{CVPR'21 Security AI Challenger: Unrestricted Adversarial Attacks on ImageNet}~\cite{chen2021unrestricted}, contestants were asked to submit adversarial examples without any access to the defense models. The dataset is a subset of ILSVRC 2012 validation set~\cite{deng2009imagenet}, which consists of 5,000 images with 5 images per class. The final score of each submission was \emph{manually} scored from two aspects: 1) image semantic and 2) quality. If the semantic of the submitted image changes (judged by human referees), then $S_s = 0$, otherwise $S_s = 1$. The image quality $S_q$ (equivalent to our reward function $\mathcal{F}_{\text{reward}}$) was quantified with five levels $S_q \in \{1,2,3,4,5\}$ by multiple human referees. The final score is given by $\sum_{i} \mathbbm{1}\{\hat{g}(\boldsymbol{x}_i^{\prime}) \neq y_i\} \times S_{s}(\boldsymbol{x}_i^{\prime})\times \frac{S_{q}(\|\boldsymbol{x}_i^{\prime} - \boldsymbol{x}_i\|)}{5}$.

We apply our method GA-DTMI-FGSM ($\eta = 0.01$) to the competition, where our entry ranked 1st place out of 1,559 teams. In the adversarial competition, our training and validation models are both an ensemble of {eight} high-performance models. We report the final score and average image quality level (equivalent to our average perturbation reward) in Fig.~\ref{fig:cvpr_1st}. It shows that our method outperforms other approaches by a large margin. In particular, we surpass the runner-up submissions by 4.59\% and 23.91\% in terms of final score and average image quality level, respectively.

\subsection{{Transferable Unrestricted Adversarial Examples}}

Most of current defenses can be easily broken by unseen attacks in a white-box manner. Adversarial training against multiple $\ell_p$-norm attacks~\cite{NEURIPS2019_5d4ae76f} solved this issue partially, however, at the cost of robustness against single $\ell_p$-norm attack. Laidlaw \etal~\cite{laidlaw2021perceptual} integrated adversarial training with Learned Perceptual Image Patch Similarity (LPIPS)~\cite{zhang2018unreasonable}, aiming to improve robustness against perturbations that were unseen during training. However, the proposed attack~\cite{laidlaw2021perceptual}, similar to other unrestricted attacks~\cite{laidlaw2019functional,engstrom2019exploring}, suffers from weaker transferability to the target model. In practice, attackers typically have no information about the defense models and the defenders do not have the ground truth to make pixel-level comparison (perturbation can be large as long as the generated adversarial examples are semantic-preserving). Therefore, we propose to benchmark classification models on ImageNet under \emph{transfer-based unrestricted attacks}.


\medskip
\noindent\textbf{Implementation Details.} For the ReColor attack~\cite{laidlaw2019functional}, we set $\varepsilon=1.0$ and iterations $T = 100$ which achieves 89.7\% attack success rate on training model $f$ (the same as Sec.~\ref{balance}) and 9.1\% transfer success rate on the test model Inception-ResNet-v2. For FSA attack~\cite{Xu_Tao_Cheng_Zhang_2021}, we set $\varepsilon=3.0$ and $T = 500$ which achieves 54.2\% attack success rate and 12.4\% transfer success rate on the same training and test models (Note that our method GA-DMI-FSA achieves 95.5\% attack success rate and 58.3\% transfer success rate, indicating that the input diversity and momentum techniques in Eq.~\eqref{DMI-FSA} boost both the attacking ability and transferability.). Besides six adversarially trained and two high-performance classification models, we select two state-of-the-art models on the  ImageNet-R dataset~\cite{hendrycks2021many}. 

\begin{figure*}[t]
  \centering
  \renewcommand{\arraystretch}{0.0}
\setlength{\tabcolsep}{2.5pt}
\newcommand{\shiftleft}[2]{\makebox[0pt][r]{\makebox[#1][l]{#2}}}
\newcommand{\imlabel}[2]{\includegraphics[width=0.104\linewidth]{#1}%
\raisebox{44pt}{\shiftleft{52pt}{\makebox[-2pt][l]{\scriptsize #2}}}}

\begin{tabular}{cc@{}cc@{}cc@{}cc@{}c}
Benign & \multicolumn{2}{c}{$\text{GA}_{\text{DTMI-FGSM}}$} & \multicolumn{2}{c}{ReColor} & \multicolumn{2}{c}{FSA} & \multicolumn{2}{c}{$\text{GA}_{\text{DMI-FSA}}$} \\
\imlabel{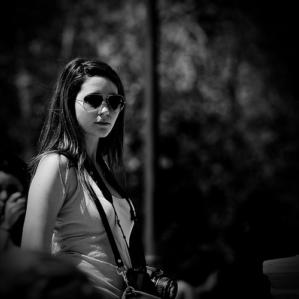}{\transparent{0.8}\colorbox{black}{\textcolor{white}{836}}} & 
\imlabel{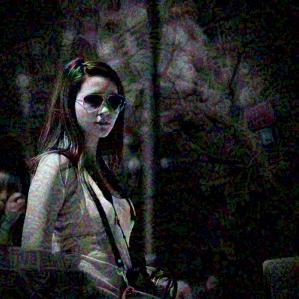}{\transparent{0.8}\colorbox{black}{\textcolor{white}{776}}} & 
\imlabel{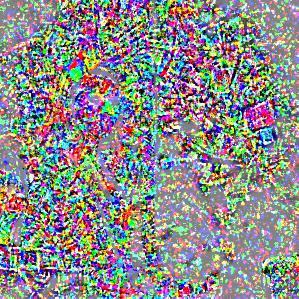}{\transparent{0.8}\colorbox{black}{\textcolor{white}{\tiny{$\varepsilon=20$}}}} & 
\imlabel{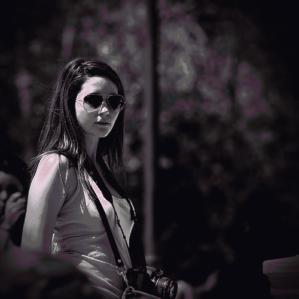}{\transparent{0.8}\colorbox{black}{\textcolor{white}{836}}} & 
\imlabel{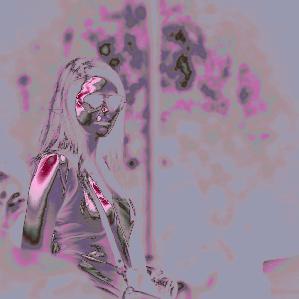}{} & 
\imlabel{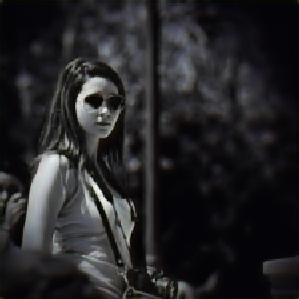}{\transparent{0.8}\colorbox{black}{\textcolor{white}{776}}} & 
\imlabel{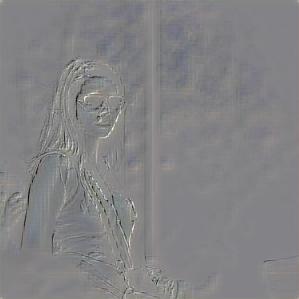}{} & 
\imlabel{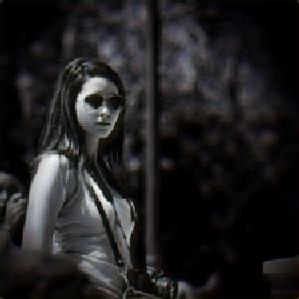}{\transparent{0.8}\colorbox{black}{\textcolor{white}{776}}} & 
\imlabel{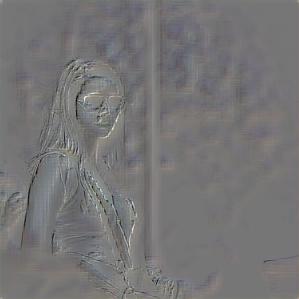}{\transparent{0.8}\colorbox{black}{\textcolor{white}{\tiny{$\varepsilon=1.28$}}}} \\
\imlabel{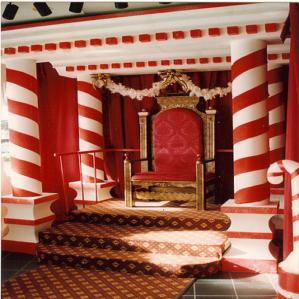}{\transparent{0.8}\colorbox{black}{\textcolor{white}{857}}} & 
\imlabel{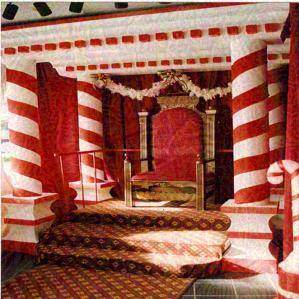}{\transparent{0.8}\colorbox{black}{\textcolor{white}{564}}} & 
\imlabel{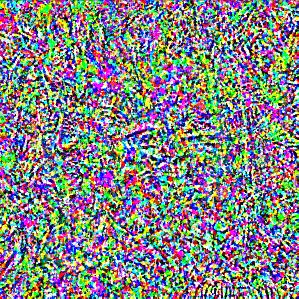}{\transparent{0.8}\colorbox{black}{\textcolor{white}{\tiny{$\varepsilon=12$}}}} & 
\imlabel{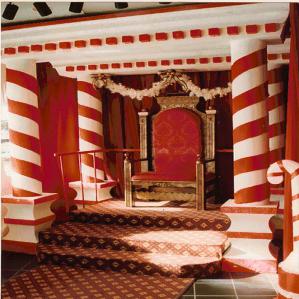}{\transparent{0.8}\colorbox{black}{\textcolor{white}{857}}} & 
\imlabel{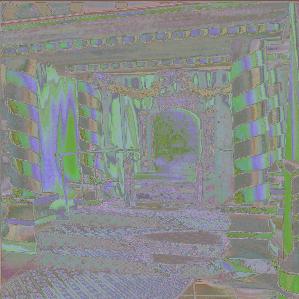}{} & 
\imlabel{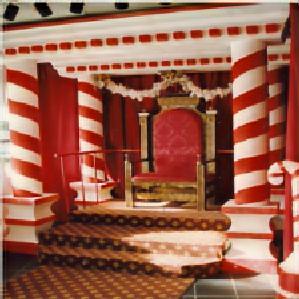}{\transparent{0.8}\colorbox{black}{\textcolor{white}{857}}} & 
\imlabel{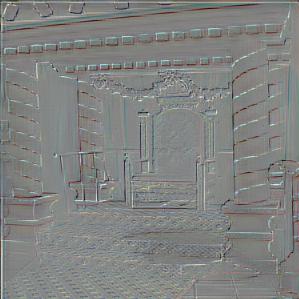}{} & 
\imlabel{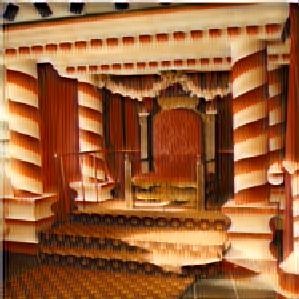}{\transparent{0.8}\colorbox{black}{\textcolor{white}{556}}} & 
\imlabel{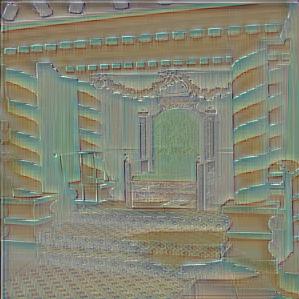}{\transparent{0.8}\colorbox{black}{\textcolor{white}{\tiny{$\varepsilon=1.65$}}}} \\
\imlabel{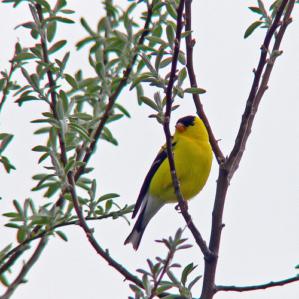}{\transparent{0.8}\colorbox{black}{\textcolor{white}{11}}} & 
\imlabel{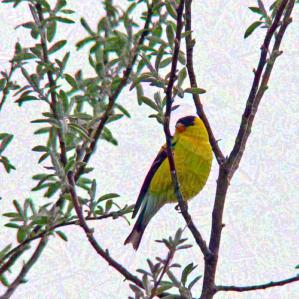}{\transparent{0.8}\colorbox{black}{\textcolor{white}{11}}} & 
\imlabel{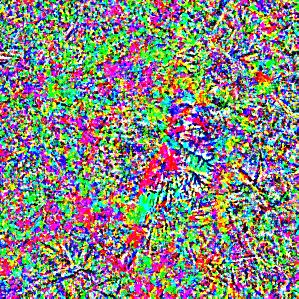}{\transparent{0.8}\colorbox{black}{\textcolor{white}{\tiny{$\varepsilon=12$}}}} & 
\imlabel{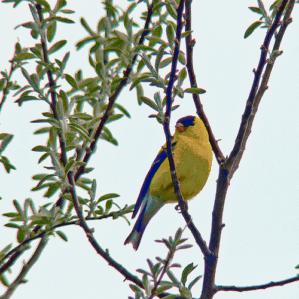}{\transparent{0.8}\colorbox{black}{\textcolor{white}{11}}} & 
\imlabel{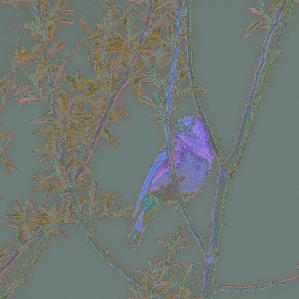}{} & 
\imlabel{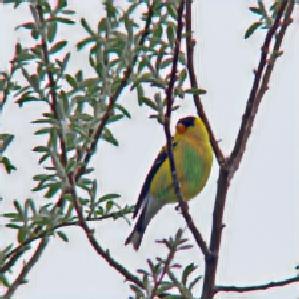}{\transparent{0.8}\colorbox{black}{\textcolor{white}{11}}} & 
\imlabel{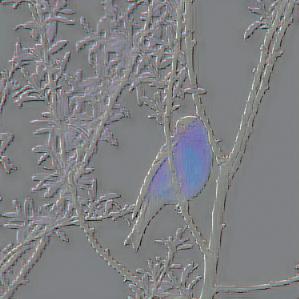}{} & 
\imlabel{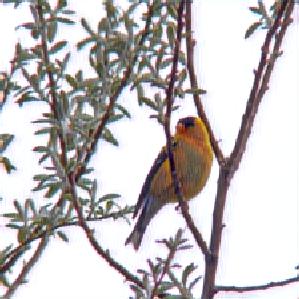}{\transparent{0.8}\colorbox{black}{\textcolor{white}{12}}} & 
\imlabel{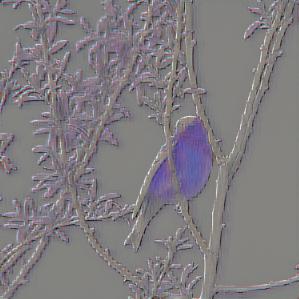}{\transparent{0.8}\colorbox{black}{\textcolor{white}{\tiny{$\varepsilon=1.65$}}}} \\
\imlabel{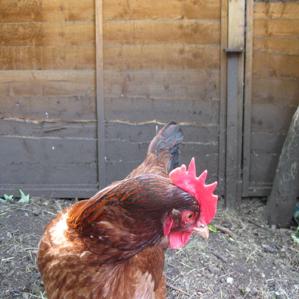}{\transparent{0.8}\colorbox{black}{\textcolor{white}{8}}} & 
\imlabel{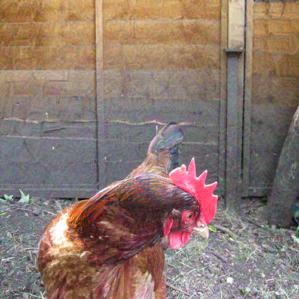}{\transparent{0.8}\colorbox{black}{\textcolor{white}{8}}} & 
\imlabel{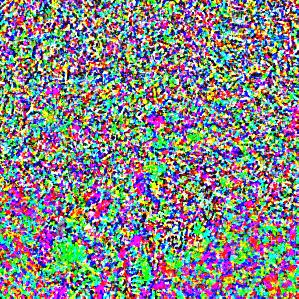}{\transparent{0.8}\colorbox{black}{\textcolor{white}{\tiny{$\varepsilon=8$}}}} & 
\imlabel{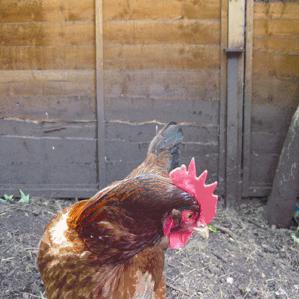}{\transparent{0.8}\colorbox{black}{\textcolor{white}{8}}} & 
\imlabel{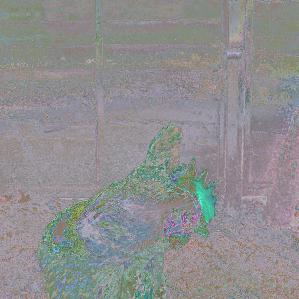}{} & 
\imlabel{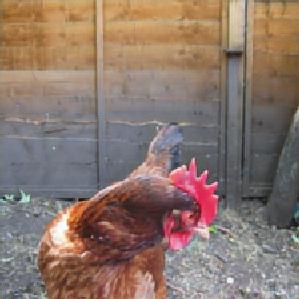}{\transparent{0.8}\colorbox{black}{\textcolor{white}{8}}} & 
\imlabel{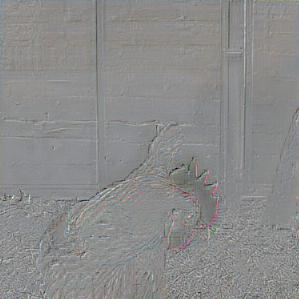}{} & 
\imlabel{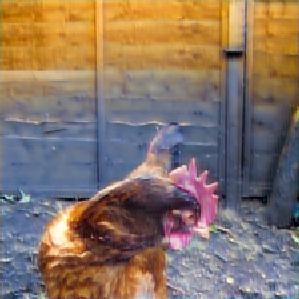}{\transparent{0.8}\colorbox{black}{\textcolor{white}{344}}} & 
\imlabel{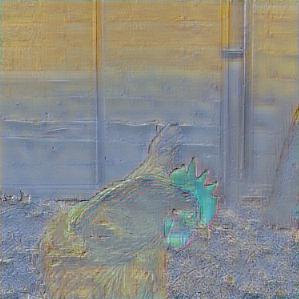}{\transparent{0.8}\colorbox{black}{\textcolor{white}{\tiny{$\varepsilon=2.72$}}}} \\
 \end{tabular}

  \caption{\textbf{Visualization of transfer attack results on Resnext101-DenoiseAll}~\cite{xie2019feature}. For each image, we print its predicted label on model Resnext101-DenoiseAll in the upper left corner. For each transfer-based adversarial example, we present the perturbation on its right. For each perturbation crafted via our geometry-aware framework, we print its perturbation budget in the upper left corner. Although the transfer-based $\ell_{\infty}$-norm attack GA-DTMI-FGSM is able to fool the defense test model to a certain extent, the generated perturbations can be ``human-perceptible" (the first and third rows of $\text{GA}_{\text{DTMI-FGSM}}$). Besides, the other two unrestricted attacks suffer from weaker transferability when compared to our method $\text{GA}_{\text{DMI-FSA}}$, which adjusts the images' color and texture that ImageNet-trained CNNs might be biased to~\cite{geirhos2018imagenet}.}
  \label{fig:attack_adv}
\end{figure*}

\begin{figure}[bh]
  \centering
  \renewcommand{\arraystretch}{0.0}
\setlength{\tabcolsep}{2.5pt}
\newcommand{\shiftleft}[2]{\makebox[0pt][r]{\makebox[#1][l]{#2}}}
\newcommand{\imlabel}[2]{\includegraphics[width=0.104\linewidth]{#1}%
\raisebox{44pt}{\shiftleft{52pt}{\makebox[-2pt][l]{\scriptsize #2}}}}

\begin{tabular}{cc@{}cc@{}cc@{}cc@{}c}
Benign & \multicolumn{2}{c}{$\text{GA}_{\text{DTMI-FGSM}}$} & \multicolumn{2}{c}{ReColor} & \multicolumn{2}{c}{FSA} & \multicolumn{2}{c}{$\text{GA}_{\text{DMI-FSA}}$} \\
\imlabel{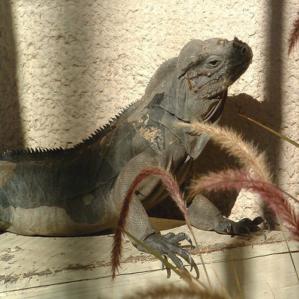}{\transparent{0.8}\colorbox{black}{\textcolor{white}{39}}} & 
\imlabel{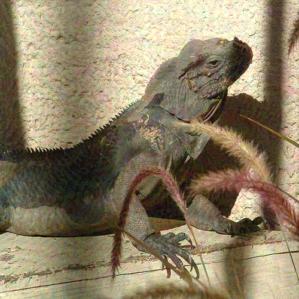}{\transparent{0.8}\colorbox{black}{\textcolor{white}{39}}} & 
\imlabel{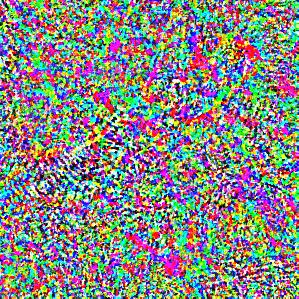}{\transparent{0.8}\colorbox{black}{\textcolor{white}{\tiny{$\varepsilon=8$}}}} & 
\imlabel{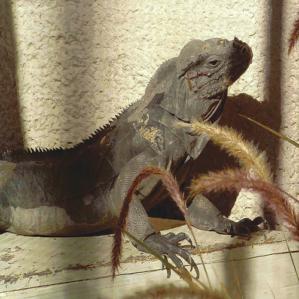}{\transparent{0.8}\colorbox{black}{\textcolor{white}{39}}} & 
\imlabel{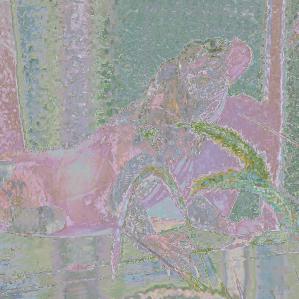}{} & 
\imlabel{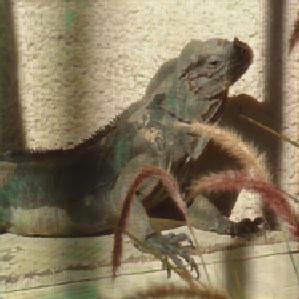}{\transparent{0.8}\colorbox{black}{\textcolor{white}{39}}} & 
\imlabel{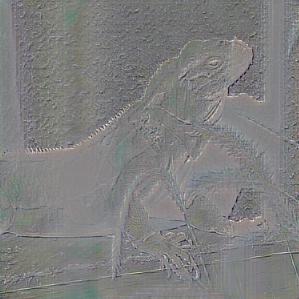}{} & 
\imlabel{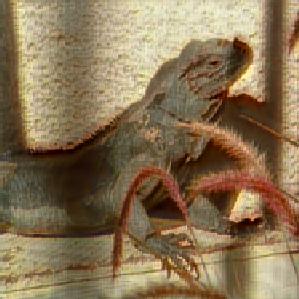}{\transparent{0.8}\colorbox{black}{\textcolor{white}{191}}} & 
\imlabel{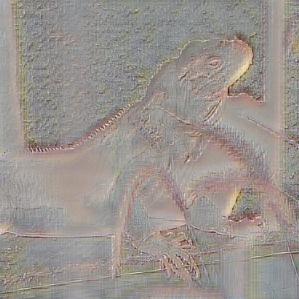}{\transparent{0.8}\colorbox{black}{\textcolor{white}{\tiny{$\varepsilon=2.12$}}}} \\
\imlabel{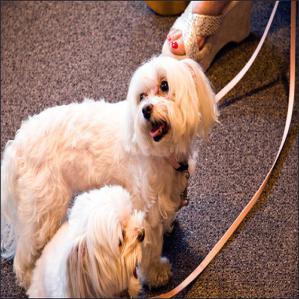}{\transparent{0.8}\colorbox{black}{\textcolor{white}{153}}} & 
\imlabel{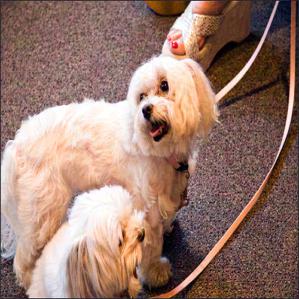}{\transparent{0.8}\colorbox{black}{\textcolor{white}{153}}} & 
\imlabel{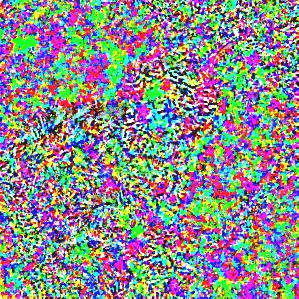}{\transparent{0.8}\colorbox{black}{\textcolor{white}{\tiny{$\varepsilon=4$}}}} & 
\imlabel{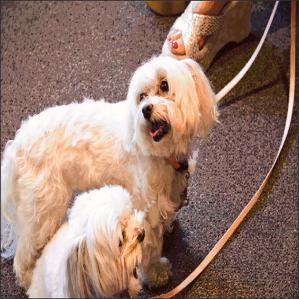}{\transparent{0.8}\colorbox{black}{\textcolor{white}{153}}} & 
\imlabel{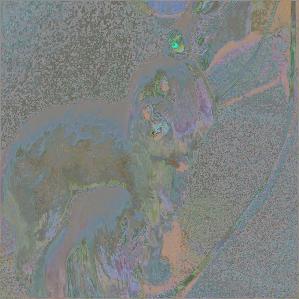}{} & 
\imlabel{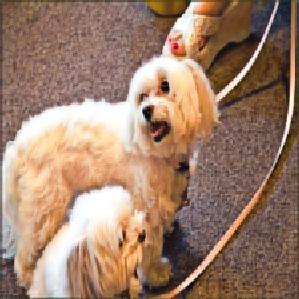}{\transparent{0.8}\colorbox{black}{\textcolor{white}{153}}} & 
\imlabel{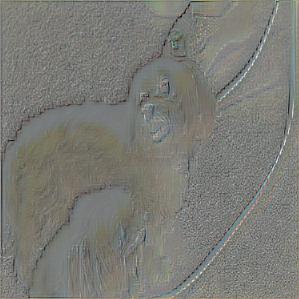}{} & 
\imlabel{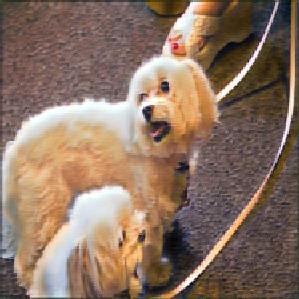}{\transparent{0.8}\colorbox{black}{\textcolor{white}{368}}} & 
\imlabel{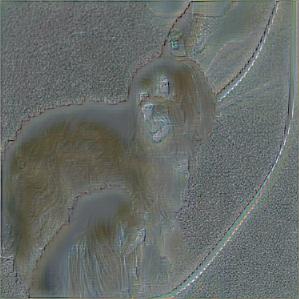}{\transparent{0.8}\colorbox{black}{\textcolor{white}{\tiny{$\varepsilon=1.65$}}}} \\
\imlabel{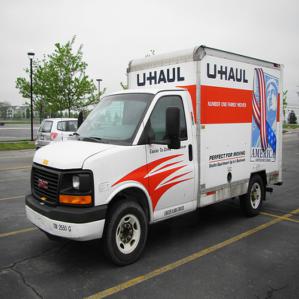}{\transparent{0.8}\colorbox{black}{\textcolor{white}{675}}} & 
\imlabel{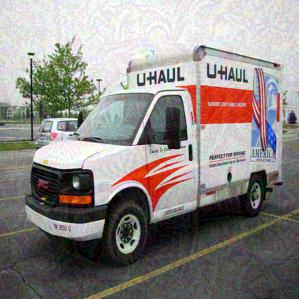}{\transparent{0.8}\colorbox{black}{\textcolor{white}{675}}} & 
\imlabel{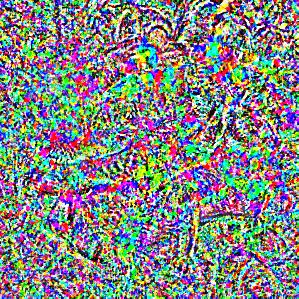}{\transparent{0.8}\colorbox{black}{\textcolor{white}{\tiny{$\varepsilon=12$}}}} & 
\imlabel{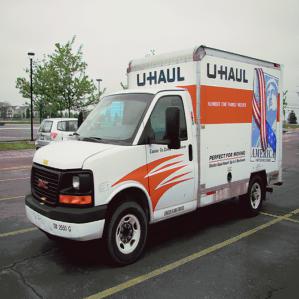}{\transparent{0.8}\colorbox{black}{\textcolor{white}{675}}} & 
\imlabel{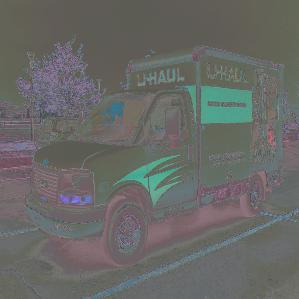}{} & 
\imlabel{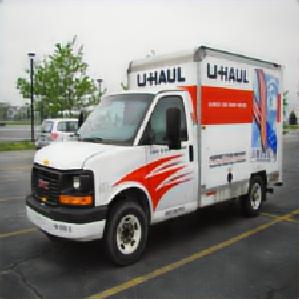}{\transparent{0.8}\colorbox{black}{\textcolor{white}{675}}} & 
\imlabel{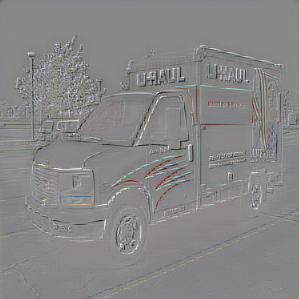}{} & 
\imlabel{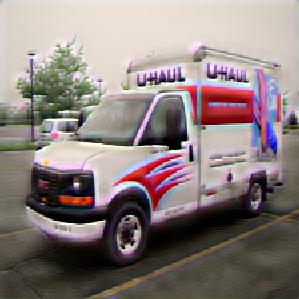}{\transparent{0.8}\colorbox{black}{\textcolor{white}{407}}} & 
\imlabel{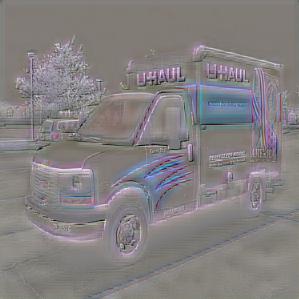}{\transparent{0.8}\colorbox{black}{\textcolor{white}{\tiny{$\varepsilon=2.12$}}}} \\
\imlabel{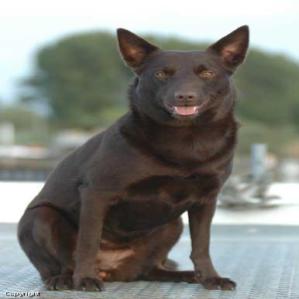}{\transparent{0.8}\colorbox{black}{\textcolor{white}{227}}} & 
\imlabel{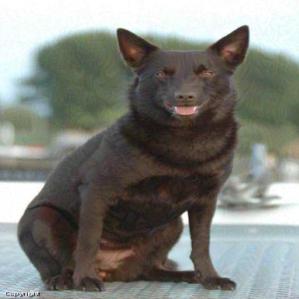}{\transparent{0.8}\colorbox{black}{\textcolor{white}{227}}} & 
\imlabel{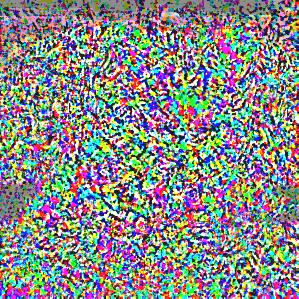}{\transparent{0.8}\colorbox{black}{\textcolor{white}{\tiny{$\varepsilon=4$}}}} & 
\imlabel{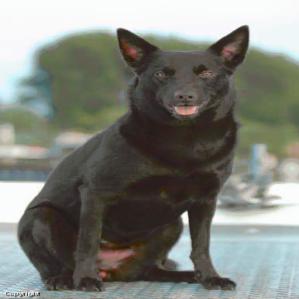}{\transparent{0.8}\colorbox{black}{\textcolor{white}{227}}} & 
\imlabel{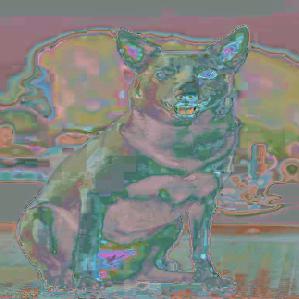}{} & 
\imlabel{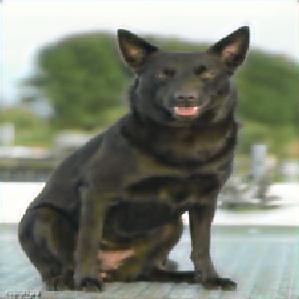}{\transparent{0.8}\colorbox{black}{\textcolor{white}{227}}} & 
\imlabel{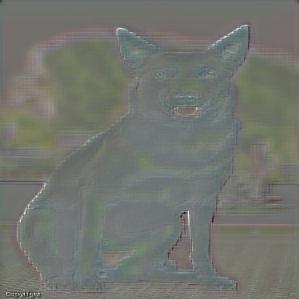}{} & 
\imlabel{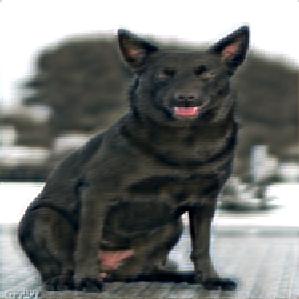}{\transparent{0.8}\colorbox{black}{\textcolor{white}{223}}} & 
\imlabel{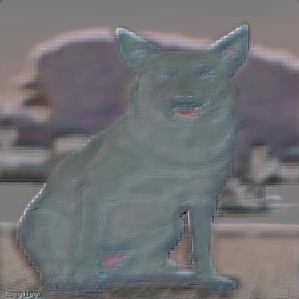}{\transparent{0.8}\colorbox{black}{\textcolor{white}{\tiny{$\varepsilon=2.12$}}}} \\
\imlabel{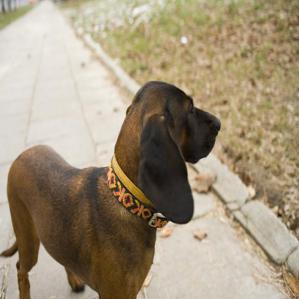}{\transparent{0.8}\colorbox{black}{\textcolor{white}{163}}} & 
\imlabel{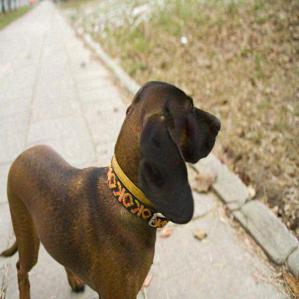}{\transparent{0.8}\colorbox{black}{\textcolor{white}{163}}} & 
\imlabel{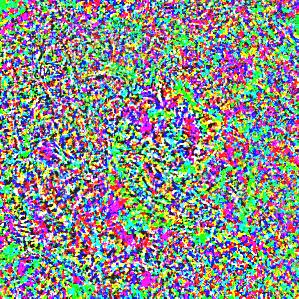}{\transparent{0.8}\colorbox{black}{\textcolor{white}{\tiny{$\varepsilon=4$}}}} & 
\imlabel{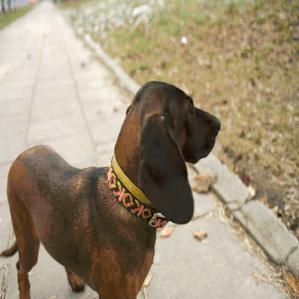}{\transparent{0.8}\colorbox{black}{\textcolor{white}{163}}} & 
\imlabel{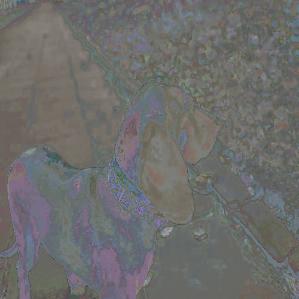}{} & 
\imlabel{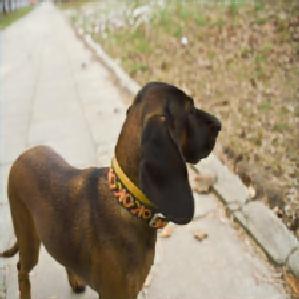}{\transparent{0.8}\colorbox{black}{\textcolor{white}{163}}} & 
\imlabel{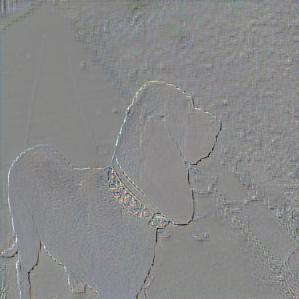}{} & 
\imlabel{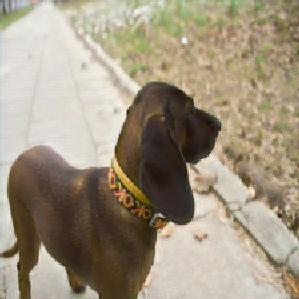}{\transparent{0.8}\colorbox{black}{\textcolor{white}{161}}} & 
\imlabel{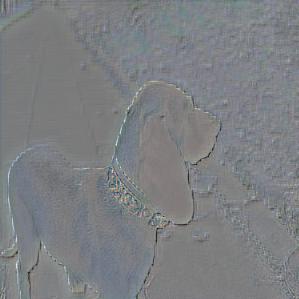}{\transparent{0.8}\colorbox{black}{\textcolor{white}{\tiny{$\varepsilon=1.28$}}}} \\
\imlabel{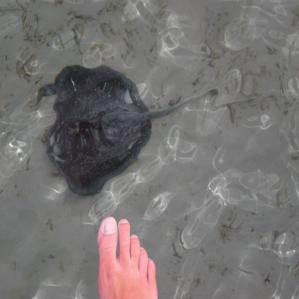}{\transparent{0.8}\colorbox{black}{\textcolor{white}{6}}} & 
\imlabel{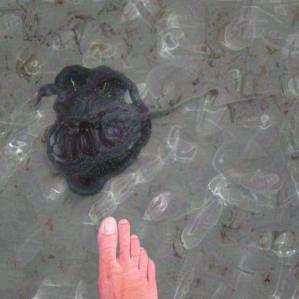}{\transparent{0.8}\colorbox{black}{\textcolor{white}{6}}} & 
\imlabel{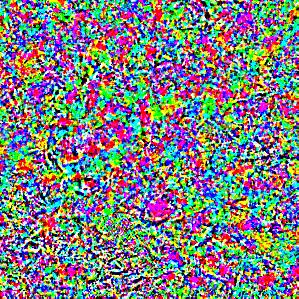}{\transparent{0.8}\colorbox{black}{\textcolor{white}{\tiny{$\varepsilon=4$}}}} & 
\imlabel{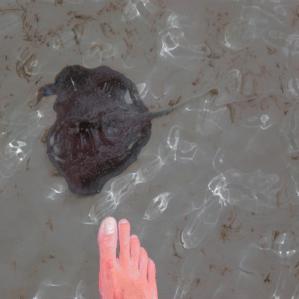}{\transparent{0.8}\colorbox{black}{\textcolor{white}{6}}} & 
\imlabel{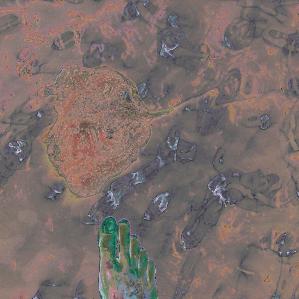}{} & 
\imlabel{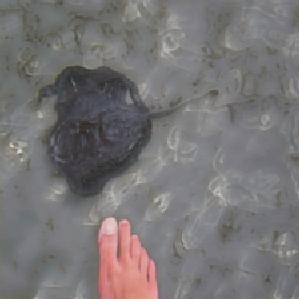}{\transparent{0.8}\colorbox{black}{\textcolor{white}{6}}} & 
\imlabel{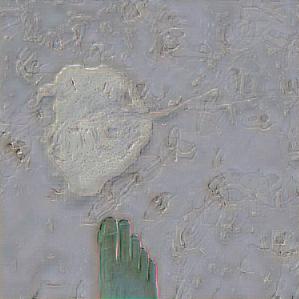}{} & 
\imlabel{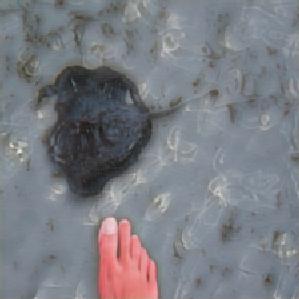}{\transparent{0.8}\colorbox{black}{\textcolor{white}{13}}} & 
\imlabel{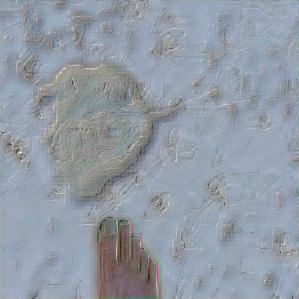}{\transparent{0.8}\colorbox{black}{\textcolor{white}{\tiny{$\varepsilon=1.28$}}}} \\
 \end{tabular}
  \caption{\textbf{Visualization of transfer attack on Resnext101-DenoiseAll}~\cite{xie2019feature}.}
  \label{fig:adv_more}
  \vspace{-4pt}
\end{figure}

\medskip
\noindent\textbf{Experimental Results.}  From Table.~\ref{benchmark}, we can conclude the following observations: a) $\text{GA}_{\text{Admix-DTI}}$ transfers better than $\text{GA}_{\text{DTMI-FGSM}}$.  b) $\ell_{\infty}$-norm transfer attack can hardly break $\ell_{\infty}$-norm adversarially trained models while the unrestricted attack (GA-DMI-FSA) reduces the accuracy of these models by a large margin (see also in Fig.~\ref{fig:attack_adv}). c) DeepAugment~\cite{hendrycks2021many}, which utilizes semantic-preserving augmentations during training, exhibits non-trivial robustness against GA-DMI-FSA attack. d) Efficientnet-l2-ns~\cite{xie2020self} performs well under all the transfer-based attacks and enjoys 86\% accuracy against GA-DMI-FSA attack, showing that the distribution of our generated adversarial examples is close to the natural examples'. Note that Efficientnet-l2-ns is also the best-performing model on the ImageNet-V2 dataset~\cite{recht2019imagenet,taori2020measuring}. We visualize part of the transfer attack results on adversarially trained Resnext101-DenoiseAll~\cite{xie2019feature} in Figs.~\ref{fig:attack_adv}, where our method GA-DMI-FSA is able to generate semantic-preserving yet transferable unrestricted adversarial examples. For more visualization results, please see Fig.~\ref{fig:adv_more} in Sec.~\ref{Visualization}.

\subsection{Ablation Studies and Discussions} \label{AS}



\begin{figure}[h]
  \begin{center}
  \includegraphics[width=0.95\columnwidth]{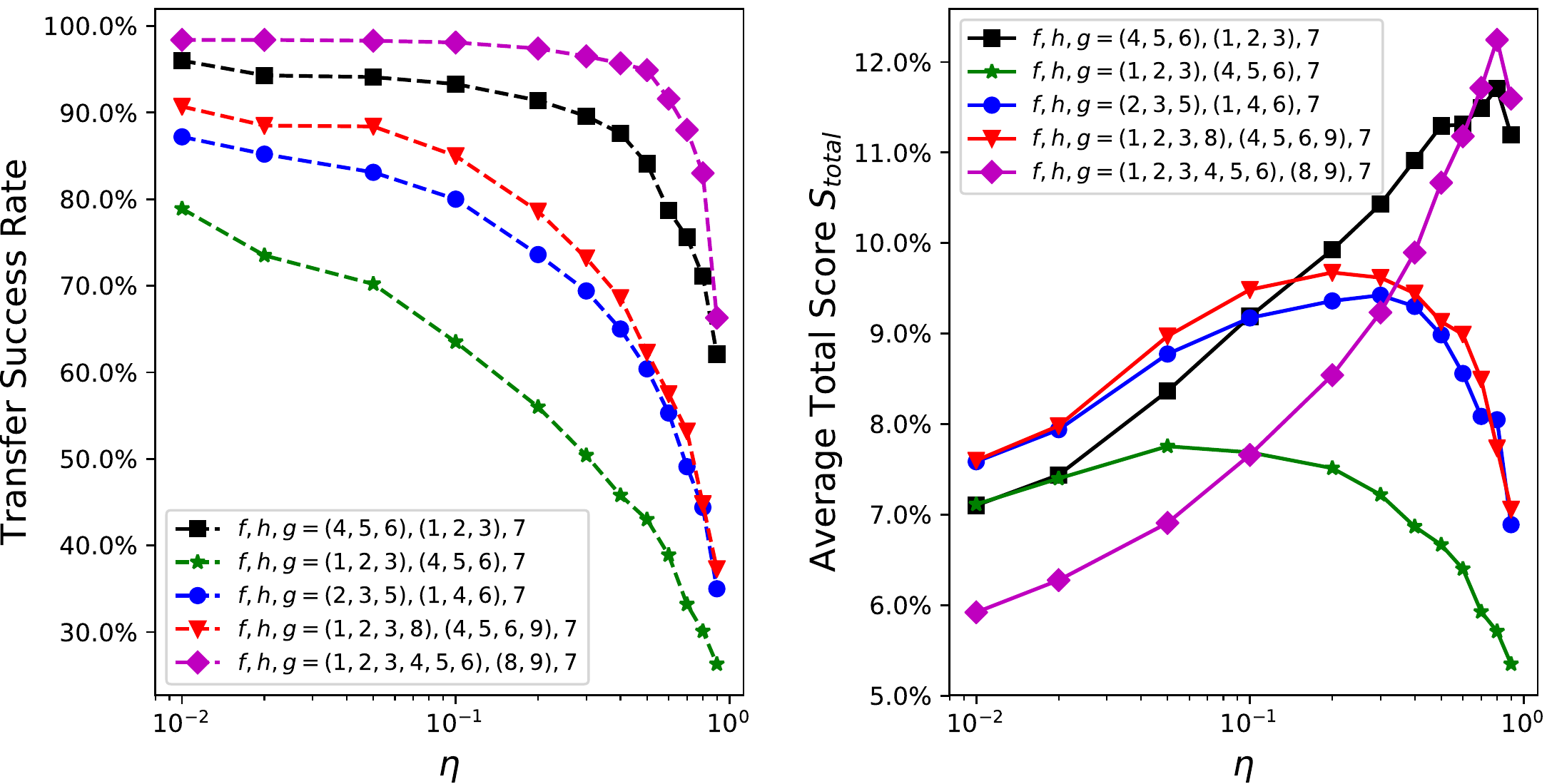}
  \end{center}
  \vspace{-10pt}
  \caption{\textbf{Comparison between various train-valid splittings.} The index in the legend corresponds to the model index in Table.~\ref{tab:models}. As the threshold $\eta$ increases, the generated adversarial examples have higher confidence (probability of the true class) on the validation model $h$, leading to a lower transfer success rate (\textbf{left}). Besides, the optimal $\eta^*$ that yields the maximum $S_{total}$ is dependent on the partition (\textbf{right}).}
  \label{fig:eta}
  \vspace{-12pt}
\end{figure}

\medskip
\noindent\textbf{The optimal $\eta$ depends on the train-valid splitting.} As declared in the implementation details in Sec.~\ref{balance}, the threshold $\eta$ is searched from a finite set ranging from 0.001 to 0.9. We now investigate how the hyper-parameter $\eta$ will affect the average total score $S_{total}$. From Fig.~\ref{fig:eta}, we observe that the optimal $\eta^*$ varies across different splittings and can be larger if the transferability from the training model $f$ to the test model $g$ is higher enough. However, as shown in Table~\ref{tab:diversity}, the improvement of our GA framework upon fixed-radius baseline (DTMI-FGSM) is \textit{stable} and independent of the splitting.

\begin{table}[h]
  \centering
  \normalsize
  \caption{\textbf{Comparison of the improvement upon fixed-radius baseline under different train-valid splittings.} The test model $g$ is  model 7 in Table~\ref{tab:models} and we select three (two of them are highlighted in Fig.~\ref{fig:represent}) representative splittings according to the partition loss $\ell_{\mathcal{G}}$. For each setting, we repeat the experiment three times and report the mean and the standard deviation (in the parenthesis).}
  \resizebox{1.0\columnwidth}{!}{
    \begin{tabular}{||ccc||ccc||}
    \hline
    $f$     & $h$ & $\ell_{\mathcal{G}}$ & Method &  $S_{APR}$ ($\uparrow$)  & $S_{total}$ ($\uparrow$) \\
    \hline
    \hline
   \multirow{2}[0]{*}{$\{4,5,6\}$}  & \multirow{2}[0]{*}{$\{1,2,3\}$}  & \multirow{2}[0]{*}{2.22} & DTMI  & 0.1080 (0.001) & 8.66\% (0.16\%) \\
    &  & & GA-DTMI & \textbf{0.1456} (6e-5) & \textbf{11.6\%} (0.21\%) \\
    \hline
    \hline
    \multirow{2}[0]{*}{$\{2,3,5\}$}  & \multirow{2}[0]{*}{$\{1,4,6\}$} & \multirow{2}[0]{*}{2.56}   & DTMI & 0.1150 (0.006) & 7.32\% (0.07\%) \\
    & & & GA-DTMI & \textbf{0.1417} (0.001) & \textbf{9.48\%} (0.03\%) \\
    \hline
    \hline
   \multirow{2}[0]{*}{$\{1,2,3\}$}  & \multirow{2}[0]{*}{$\{4,5,6\}$} & \multirow{2}[0]{*}{2.70}   & DTMI & 0.1273 (0.007) & 6.41\% (0.07\%) \\
    & & &  GA-DTMI & \textbf{0.1291} (0.006) & \textbf{8.23\%} (0.05\%) \\
    \hline
    \end{tabular}%
    }
  \label{tab:diversity}%
\end{table}%

\begin{figure*}[t]
  \centering
  \begin{subfigure}{0.328\linewidth}
    \includegraphics[width=1.0\linewidth]{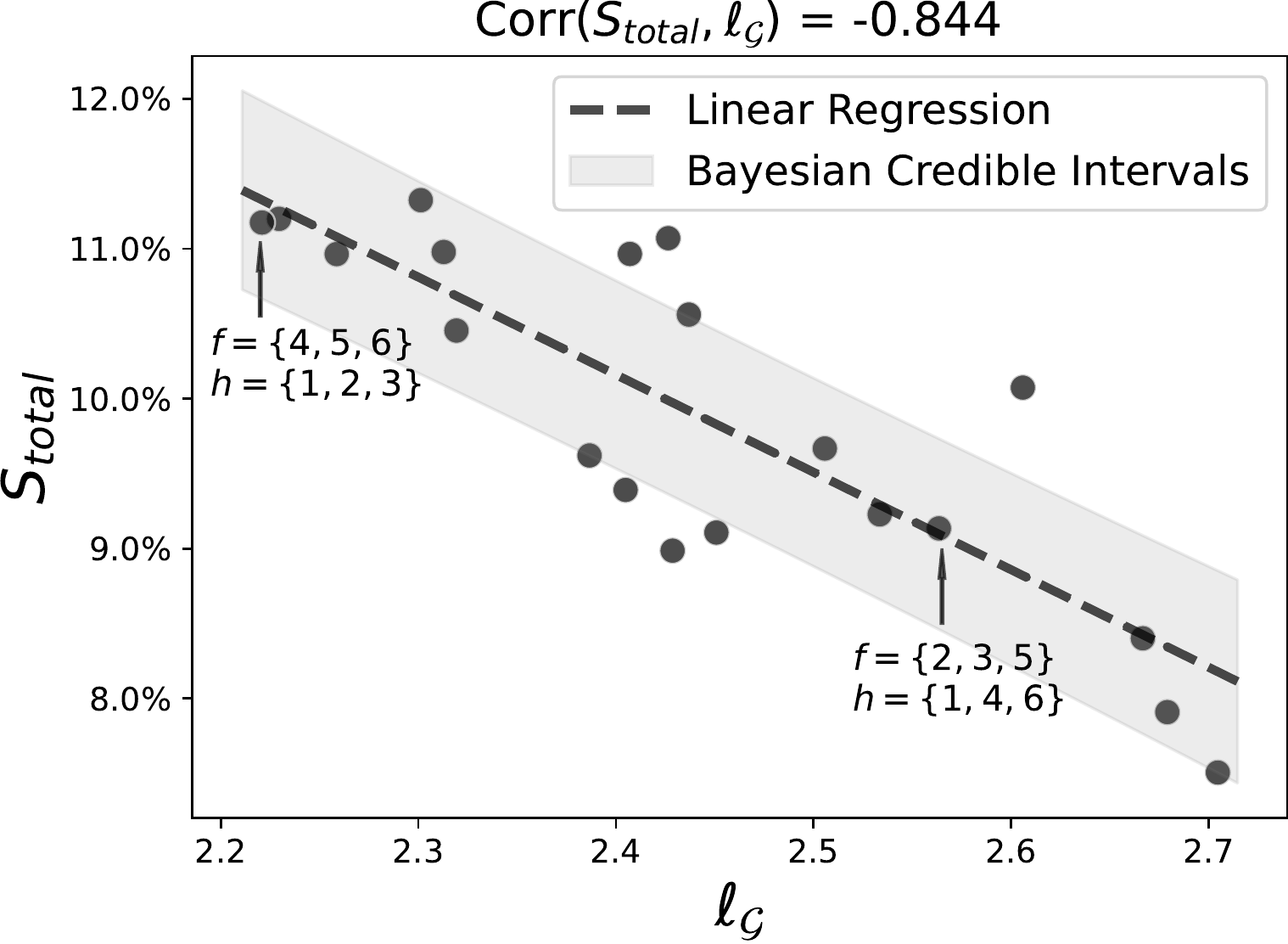}
    \caption{$f \cup h = \{1,\cdots,6\}, (n, k) = (6, 3)$.}
  \label{fig:represent}
  \end{subfigure}
  \hfill
  \begin{subfigure}{0.328\linewidth}
    \includegraphics[width=1.0\linewidth]{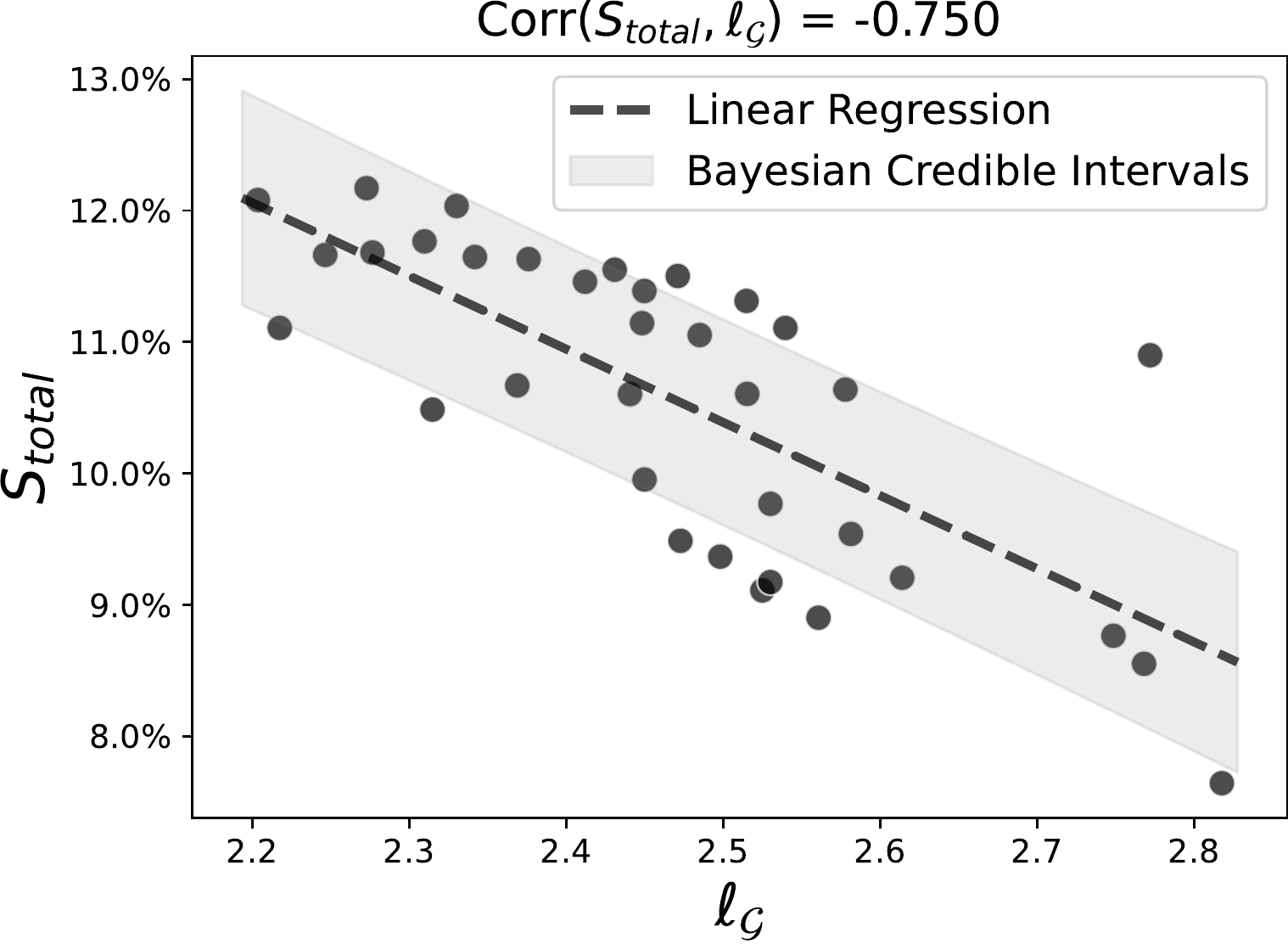}
   \caption{$f \cup h = \{0, \cdots, 6\}, (n, k) = (7, 4)$.}
  \label{fig:overfitting}
  \end{subfigure}
  \hfill
  \begin{subfigure}{0.328\linewidth}
    \includegraphics[width=1.0\linewidth]{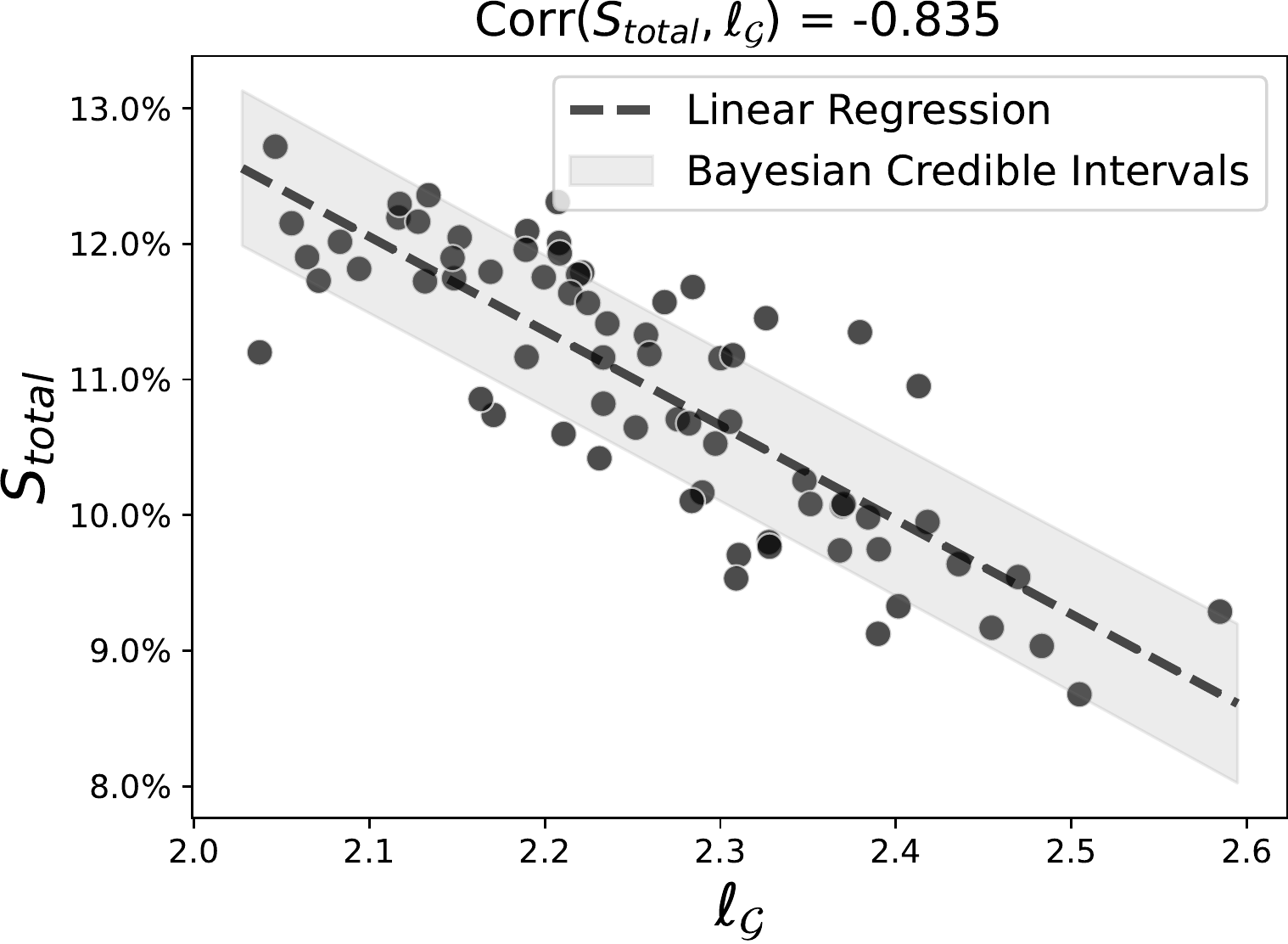}
   \caption{$f \cup h = \{0, \cdots, 6\} \cup \{8\}, (n, k) = (8, 4)$.}
  \label{fig:curve}
  \end{subfigure}
  \caption{\textbf{The effectiveness of the partition loss $\ell_{\mathcal{G}}$ under different numbers of pre-trained models.} The test model $g$ is  model 7 in Table~\ref{tab:models}. We conduct bayesian ridge regression and plot the mean of the predictive distribution as dashed lines. The Bayesian Credible Intervals range from mean - standard deviation (of the predictive distribution) to mean + standard deviation.}
  \label{number}
  \vspace{-5pt}
\end{figure*}

\begin{figure*}[t]
  \centering
  \begin{subfigure}{0.328\linewidth}
    \includegraphics[width=1.0\linewidth]{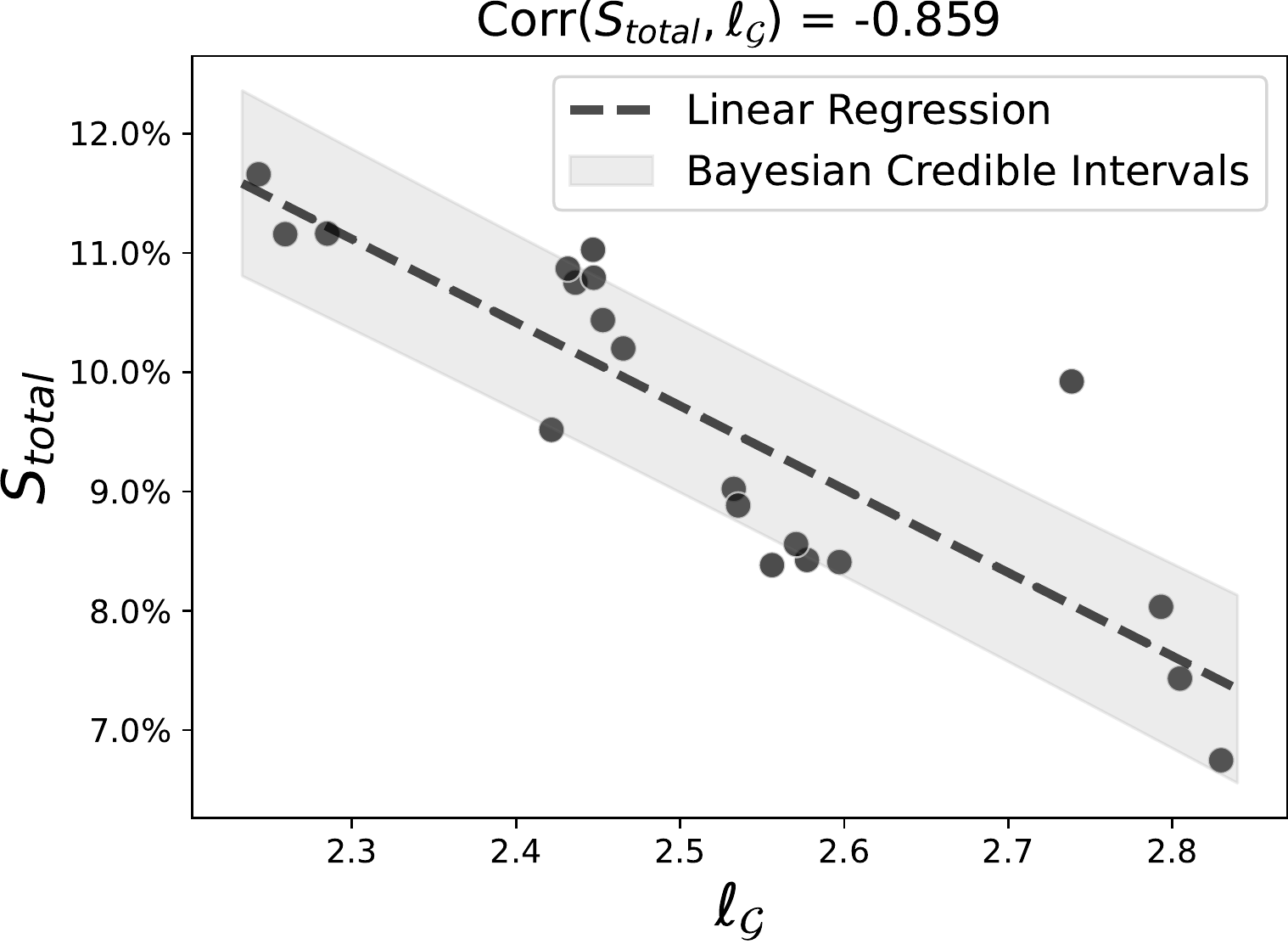}
    \caption{$f \cup h = \{0,1,\cdots,6\} \setminus \{1\}$.}
  \end{subfigure}
  \hfill
  \begin{subfigure}{0.328\linewidth}
    \includegraphics[width=1.0\linewidth]{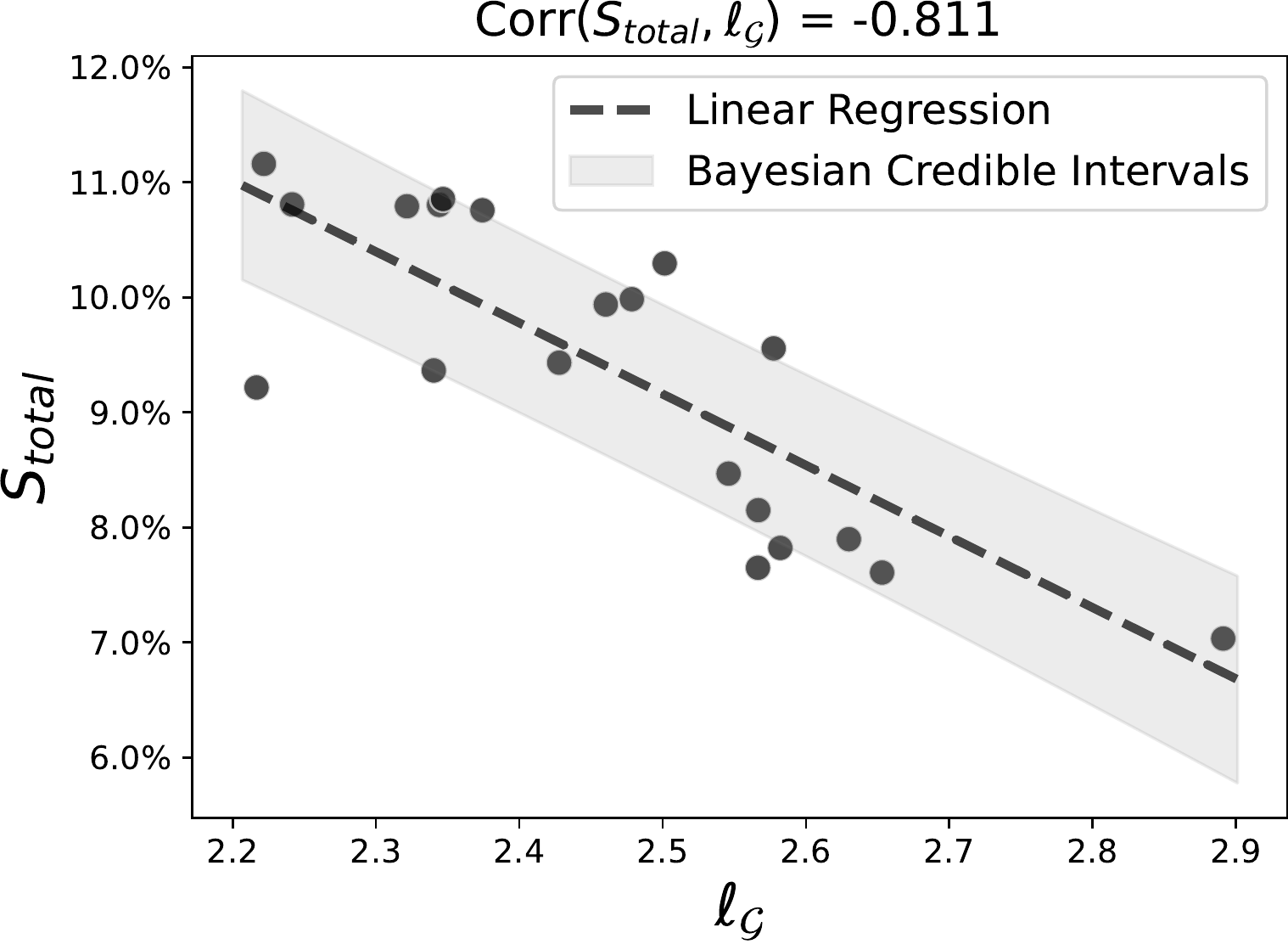}
   \caption{$f \cup h = \{0,1,\cdots,6\} \setminus \{3\}$.}
  \end{subfigure}
  \hfill
  \begin{subfigure}{0.328\linewidth}
    \includegraphics[width=1.0\linewidth]{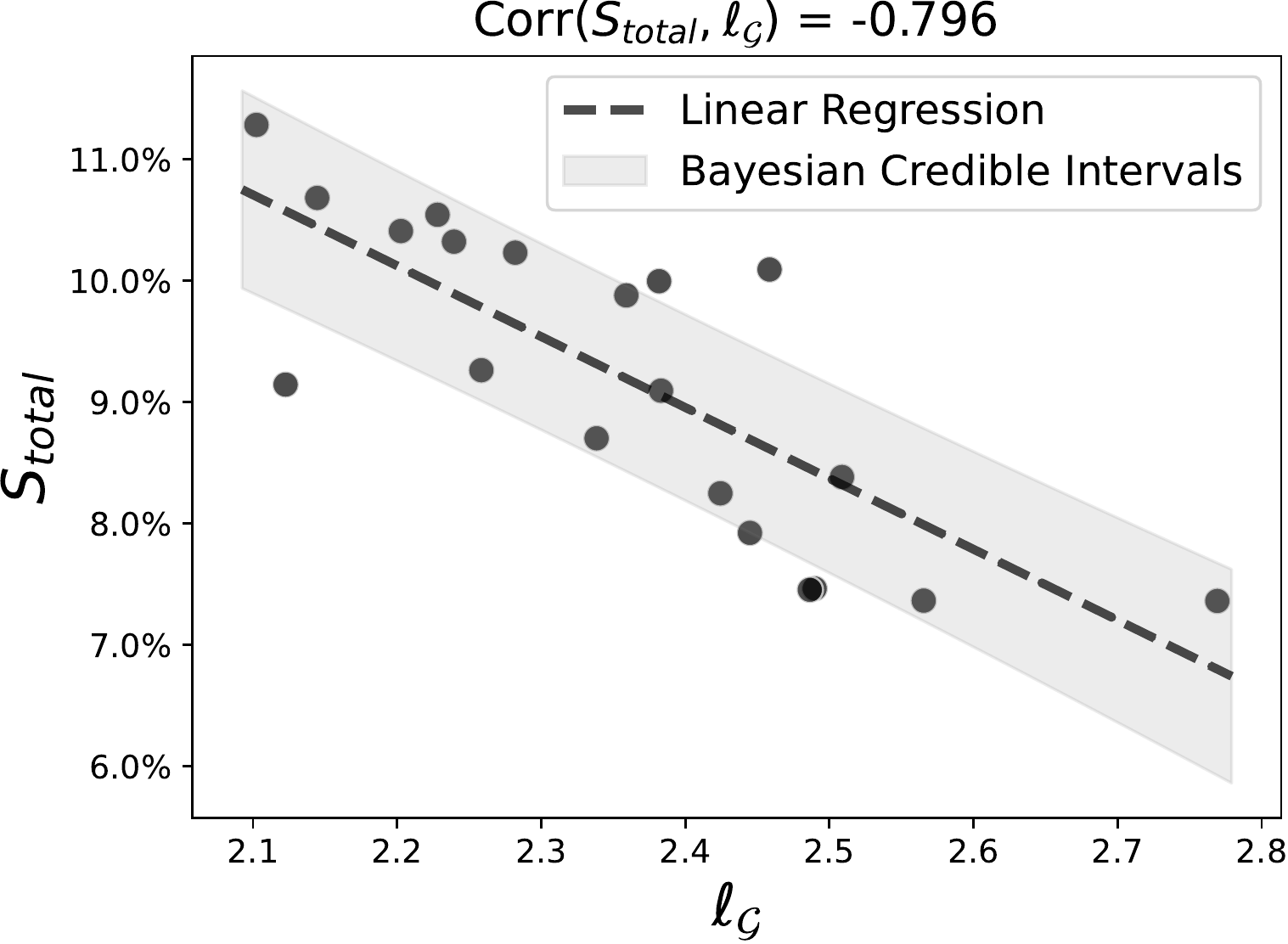}
    \caption{$f \cup h = \{0,1,\cdots,6\} \setminus \{5\}$.}
  \end{subfigure}
  \caption{\textbf{The effectiveness of the partition loss $\ell_{\mathcal{G}}$ under different kinds of pre-trained models.} The test model $g$ is  model 7 in Table~\ref{tab:models}. We simulate different kinds of pre-trained models by removing one model from the fixed set such that $(n, k) = (6, 3)$.}
  \label{kind}
  \vspace{-5pt}
\end{figure*}

\medskip
\noindent\textbf{The effectiveness of $\ell_{\mathcal{G}}$ under different numbers of pre-trained models and different kinds of pre-trained models.} To investigate the robustness of the proposed partition loss $\ell_{\mathcal{G}}$ under different settings, we carefully design controlled experiments in Figs.~\ref{number} and \ref{kind}. There are total $C_n^k$ kinds of partitions when selecting $k$ training models from $n$ pre-trained models. Given $n$ and $k$, we run our method GA-DTMI-FGSM for all the train-valid splittings and obtain a scatter plot with $C_n^k$ points. We observe a strong negative correlation between the partition loss $\ell_{\mathcal{G}}$ and the average total score $S_{total}$. For example, the average Pearson correlation coefficient over the six scatterplots in Figs.~\ref{number} and \ref{kind} is around -0.82. Moreover, the negative correlation is significant and consistent across different numbers of pre-trained models (see Fig.~\ref{number}) and different kinds of pre-trained models (see Fig.~\ref{kind}).

\section{Conclusion}
\label{sec:conclu}

In this work, we propose a geometry-aware framework, where fixed-radius methods can be integrated to generate transferable {unrestricted} adversarial examples with minimum changes. Under $\ell_{\infty}$-norm setting, our framework could improve the imperceptibility of the crafted adversarial examples by a large margin without the decrease of transfer success rate. Besides, we propose a transfer-based unrestricted attack by combining the white-box feature space attack with transfer-based $\ell_{\infty}$-norm attacks to generate semantic-preserving {yet} transferable unrestricted adversarial examples.

\medskip
\noindent\textbf{Acknowledgements.} Fangcheng Liu and Chao Zhang are supported by the National Nature Science Foundation of China under Grant 62071013 and 61671027, and National Key R\&D Program of China under Grant 2018AAA0100300. Hongyang Zhang is supported by NSERC Discovery Grant RGPIN-2022-03215, DGECR-2022-00357.

\bibliographystyle{IEEEtran}
\bibliography{references}{}

\begin{thebibliography}{10}
\providecommand{\url}[1]{#1}
\csname url@samestyle\endcsname
\providecommand{\newblock}{\relax}
\providecommand{\bibinfo}[2]{#2}
\providecommand{\BIBentrySTDinterwordspacing}{\spaceskip=0pt\relax}
\providecommand{\BIBentryALTinterwordstretchfactor}{4}
\providecommand{\BIBentryALTinterwordspacing}{\spaceskip=\fontdimen2\font plus
\BIBentryALTinterwordstretchfactor\fontdimen3\font minus
  \fontdimen4\font\relax}
\providecommand{\BIBforeignlanguage}[2]{{%
\expandafter\ifx\csname l@#1\endcsname\relax
\typeout{** WARNING: IEEEtran.bst: No hyphenation pattern has been}%
\typeout{** loaded for the language `#1'. Using the pattern for}%
\typeout{** the default language instead.}%
\else
\language=\csname l@#1\endcsname
\fi
#2}}
\providecommand{\BIBdecl}{\relax}
\BIBdecl

\bibitem{he2016deep}
K.~He, X.~Zhang, S.~Ren, and J.~Sun, ``Deep residual learning for image
  recognition,'' in \emph{CVPR}, 2016, pp. 770--778.

\bibitem{dosovitskiy2021an}
A.~Dosovitskiy, L.~Beyer, A.~Kolesnikov, D.~Weissenborn, X.~Zhai,
  T.~Unterthiner, M.~Dehghani, M.~Minderer, G.~Heigold, S.~Gelly, J.~Uszkoreit,
  and N.~Houlsby, ``An image is worth 16x16 words: Transformers for image
  recognition at scale,'' in \emph{ICLR}, 2021.

\bibitem{szegedy2013intriguing}
C.~Szegedy, W.~Zaremba, I.~Sutskever, J.~Bruna, D.~Erhan, I.~Goodfellow, and
  R.~Fergus, ``Intriguing properties of neural networks,'' in \emph{ICLR},
  2014.

\bibitem{DBLP:journals/corr/GoodfellowSS14}
I.~J. Goodfellow, J.~Shlens, and C.~Szegedy, ``Explaining and harnessing
  adversarial examples,'' in \emph{ICLR}, 2015.

\bibitem{shao2021adversarial}
R.~Shao, Z.~Shi, J.~Yi, P.-Y. Chen, and C.-J. Hsieh, ``On the adversarial
  robustness of visual transformers,'' \emph{arXiv preprint arXiv:2103.15670},
  2021.

\bibitem{Bhojanapalli_2021_ICCV}
S.~Bhojanapalli, A.~Chakrabarti, D.~Glasner, D.~Li, T.~Unterthiner, and
  A.~Veit, ``Understanding robustness of transformers for image
  classification,'' in \emph{ICCV}, October 2021, pp. 10\,231--10\,241.

\bibitem{bai2021transformers}
Y.~Bai, J.~Mei, A.~Yuille, and C.~Xie, ``Are transformers more robust than
  cnns?'' in \emph{NeurIPS}, 2021.

\bibitem{bojarski2016end}
M.~Bojarski, D.~Del~Testa, D.~Dworakowski, B.~Firner, B.~Flepp, P.~Goyal, L.~D.
  Jackel, M.~Monfort, U.~Muller, J.~Zhang \emph{et~al.}, ``End to end learning
  for self-driving cars,'' \emph{arXiv preprint arXiv:1604.07316}, 2016.

\bibitem{BMVC2015_41}
O.~M. Parkhi, A.~Vedaldi, and A.~Zisserman, ``Deep face recognition,'' in
  \emph{BMVC}.\hskip 1em plus 0.5em minus 0.4em\relax BMVA Press, September
  2015, pp. 41.1--41.12.

\bibitem{Athalye2018}
A.~Athalye, N.~Carlini, and D.~Wagner, ``{Obfuscated gradients give a false
  sense of security: Circumventing defenses to adversarial examples},'' in
  \emph{ICML}, vol.~1, 2018, pp. 436--448.

\bibitem{NEURIPS2020_11f38f8e}
F.~Tramer, N.~Carlini, W.~Brendel, and A.~Madry, ``On adaptive attacks to
  adversarial example defenses,'' in \emph{NeurIPS}, vol.~33.\hskip 1em plus
  0.5em minus 0.4em\relax Curran Associates, Inc., 2020, pp. 1633--1645.

\bibitem{papernot2016transferability}
N.~Papernot, P.~McDaniel, and I.~Goodfellow, ``Transferability in machine
  learning: from phenomena to black-box attacks using adversarial samples,''
  \emph{arXiv preprint arXiv:1605.07277}, 2016.

\bibitem{papernot2017practical}
N.~Papernot, P.~McDaniel, I.~Goodfellow, S.~Jha, Z.~B. Celik, and A.~Swami,
  ``Practical black-box attacks against machine learning,'' in
  \emph{Proceedings of the 2017 ACM on Asia conference on computer and
  communications security}, 2017, pp. 506--519.

\bibitem{liu2017delving}
Y.~Liu, X.~Chen, C.~Liu, and D.~Song, ``Delving into transferable adversarial
  examples and black-box attacks,'' in \emph{ICLR}, 2017.

\bibitem{tram2018}
F.~Tramer, A.~Kurakin, N.~Papernot, I.~Goodfellow, D.~Boneh, and P.~McDaniel,
  ``Ensemble adversarial training: Attacks and defenses,'' in \emph{ICLR},
  2018.

\bibitem{dong2018boosting}
Y.~Dong, F.~Liao, T.~Pang, H.~Su, J.~Zhu, X.~Hu, and J.~Li, ``Boosting
  adversarial attacks with momentum,'' in \emph{CVPR}, 2018, pp. 9185--9193.

\bibitem{xie2019improving}
C.~Xie, Z.~Zhang, Y.~Zhou, S.~Bai, J.~Wang, Z.~Ren, and A.~Yuille, ``Improving
  transferability of adversarial examples with input diversity,'' in
  \emph{CVPR}, 2019.

\bibitem{dong2019evading}
Y.~Dong, T.~Pang, H.~Su, and J.~Zhu, ``Evading defenses to transferable
  adversarial examples by translation-invariant attacks,'' in \emph{CVPR},
  2019, pp. 4312--4321.

\bibitem{xiao2018spatially}
C.~Xiao, J.-Y. Zhu, B.~Li, W.~He, M.~Liu, and D.~Song, ``Spatially transformed
  adversarial examples,'' in \emph{ICLR}, 2018.

\bibitem{wong2019wasserstein}
E.~Wong, F.~Schmidt, and Z.~Kolter, ``Wasserstein adversarial examples via
  projected sinkhorn iterations,'' in \emph{ICML}.\hskip 1em plus 0.5em minus
  0.4em\relax PMLR, 2019, pp. 6808--6817.

\bibitem{laidlaw2021perceptual}
C.~Laidlaw, S.~Singla, and S.~Feizi, ``Perceptual adversarial robustness:
  Defense against unseen threat models,'' in \emph{ICLR}, 2021.

\bibitem{NEURIPS2019_32508f53}
S.~Cheng, Y.~Dong, T.~Pang, H.~Su, and J.~Zhu, ``Improving black-box
  adversarial attacks with a transfer-based prior,'' in \emph{NeurIPS},
  vol.~32, 2019.

\bibitem{katzir2021s}
Z.~Katzir and Y.~Elovici, ``Who's afraid of adversarial transferability?''
  \emph{arXiv preprint arXiv:2105.00433}, 2021.

\bibitem{chen2021unrestricted}
Y.~Chen, X.~Mao, Y.~He, H.~Xue, C.~Li, Y.~Dong, Q.-A. Fu, X.~Yang, W.~Xiang,
  T.~Pang \emph{et~al.}, ``Unrestricted adversarial attacks on {ImageNet}
  competition,'' \emph{arXiv preprint arXiv:2110.09903}, 2021.

\bibitem{Xu_Tao_Cheng_Zhang_2021}
Q.~Xu, G.~Tao, S.~Cheng, and X.~Zhang, ``Towards feature space adversarial
  attack by style perturbation,'' \emph{AAAI}, vol.~35, no.~12, pp.
  10\,523--10\,531, May 2021.

\bibitem{madry2017towards}
A.~Madry, A.~Makelov, L.~Schmidt, D.~Tsipras, and A.~Vladu, ``Towards deep
  learning models resistant to adversarial attacks,'' in \emph{ICLR}, 2018.

\bibitem{Kurakin2019}
A.~Kurakin, I.~J. Goodfellow, and S.~Bengio, ``{Adversarial examples in the
  physical world},'' \emph{ICLR 2017 - Workshop Track Proceedings}, no.~c, pp.
  1--14, 2019.

\bibitem{moosavi2016deepfool}
S.-M. Moosavi-Dezfooli, A.~Fawzi, and P.~Frossard, ``Deepfool: a simple and
  accurate method to fool deep neural networks,'' in \emph{CVPR}, 2016, pp.
  2574--2582.

\bibitem{carlini2017towards}
N.~Carlini and D.~Wagner, ``Towards evaluating the robustness of neural
  networks,'' in \emph{2017 IEEE symposium on security and privacy (sp)}.\hskip
  1em plus 0.5em minus 0.4em\relax IEEE, 2017, pp. 39--57.

\bibitem{croce2020minimally}
F.~Croce and M.~Hein, ``Minimally distorted adversarial examples with a fast
  adaptive boundary attack,'' in \emph{ICML}.\hskip 1em plus 0.5em minus
  0.4em\relax PMLR, 2020, pp. 2196--2205.

\bibitem{chen2017zoo}
P.-Y. Chen, H.~Zhang, Y.~Sharma, J.~Yi, and C.-J. Hsieh, ``Zoo: Zeroth order
  optimization based black-box attacks to deep neural networks without training
  substitute models,'' in \emph{Proceedings of the 10th ACM workshop on
  artificial intelligence and security}, 2017, pp. 15--26.

\bibitem{andriushchenko2020square}
M.~Andriushchenko, F.~Croce, N.~Flammarion, and M.~Hein, ``Square attack: {A}
  query-efficient black-box adversarial attack via random search,'' in
  \emph{ECCV}, 2020, pp. 484--501.

\bibitem{brendel2018decisionbased}
W.~Brendel, J.~Rauber, and M.~Bethge, ``Decision-based adversarial attacks:
  Reliable attacks against black-box machine learning models,'' in \emph{ICLR},
  2018.

\bibitem{cheng2018queryefficient}
M.~Cheng, T.~Le, P.-Y. Chen, H.~Zhang, J.~Yi, and C.-J. Hsieh,
  ``Query-efficient hard-label black-box attack: An optimization-based
  approach,'' in \emph{ICLR}, 2019.

\bibitem{willmott2021you}
D.~Willmott, A.~K. Sahu, F.~Sheikholeslami, F.~Condessa, and Z.~Kolter, ``You
  only query once: Effective black box adversarial attacks with minimal
  repeated queries,'' \emph{arXiv preprint arXiv:2102.00029}, 2021.

\bibitem{Lin2020Nesterov}
J.~Lin, C.~Song, K.~He, L.~Wang, and J.~E. Hopcroft, ``Nesterov accelerated
  gradient and scale invariance for adversarial attacks,'' in
  \emph{International Conference on Learning Representations}, 2020.

\bibitem{Wu2020Skip}
D.~Wu, Y.~Wang, S.-T. Xia, J.~Bailey, and X.~Ma, ``Skip connections matter: On
  the transferability of adversarial examples generated with resnets,'' in
  \emph{International Conference on Learning Representations}, 2020.

\bibitem{wang2021admix}
X.~Wang, X.~He, J.~Wang, and K.~He, ``Admix: Enhancing the transferability of
  adversarial attacks,'' in \emph{Proceedings of the IEEE/CVF International
  Conference on Computer Vision}, 2021, pp. 16\,158--16\,167.

\bibitem{wang2021a}
X.~Wang, J.~Ren, S.~Lin, X.~Zhu, Y.~Wang, and Q.~Zhang, ``A unified approach to
  interpreting and boosting adversarial transferability,'' in
  \emph{International Conference on Learning Representations}, 2021.

\bibitem{johnson2016perceptual}
J.~Johnson, A.~Alahi, and L.~Fei-Fei, ``Perceptual losses for real-time style
  transfer and super-resolution,'' in \emph{ECCV}.\hskip 1em plus 0.5em minus
  0.4em\relax Springer, 2016, pp. 694--711.

\bibitem{isola2017image}
P.~Isola, J.-Y. Zhu, T.~Zhou, and A.~A. Efros, ``Image-to-image translation
  with conditional adversarial networks,'' in \emph{CVPR}, 2017, pp.
  1125--1134.

\bibitem{brown2018unrestricted}
T.~B. Brown, N.~Carlini, C.~Zhang, C.~Olsson, P.~Christiano, and I.~Goodfellow,
  ``Unrestricted adversarial examples,'' \emph{arXiv preprint
  arXiv:1809.08352}, 2018.

\bibitem{alaifari2018adef}
R.~Alaifari, G.~S. Alberti, and T.~Gauksson, ``{AD}ef: an iterative algorithm
  to construct adversarial deformations,'' in \emph{ICLR}, 2019.

\bibitem{engstrom2019exploring}
L.~Engstrom, B.~Tran, D.~Tsipras, L.~Schmidt, and A.~Madry, ``Exploring the
  landscape of spatial robustness,'' in \emph{ICML}.\hskip 1em plus 0.5em minus
  0.4em\relax PMLR, 2019, pp. 1802--1811.

\bibitem{hosseini2018semantic}
H.~Hosseini and R.~Poovendran, ``Semantic adversarial examples,'' in
  \emph{Proceedings of the IEEE Conference on Computer Vision and Pattern
  Recognition Workshops}, 2018, pp. 1614--1619.

\bibitem{laidlaw2019functional}
C.~Laidlaw and S.~Feizi, ``Functional adversarial attacks,'' in \emph{NeurIPS},
  2019.

\bibitem{zhao2020towards}
Z.~Zhao, Z.~Liu, and M.~Larson, ``Towards large yet imperceptible adversarial
  image perturbations with perceptual color distance,'' in \emph{CVPR}, 2020,
  pp. 1039--1048.

\bibitem{conf/bmvc/ZhaoLL20}
Z.~Zhao, Z.~Liu, and M.~A. Larson, ``Adversarial color enhancement: Generating
  unrestricted adversarial images by optimizing a color filter,'' in
  \emph{BMVC}, 2020.

\bibitem{shamsabadi2020colorfool}
A.~S. Shamsabadi, R.~Sanchez-Matilla, and A.~Cavallaro, ``Colorfool: Semantic
  adversarial colorization,'' in \emph{CVPR}, 2020, pp. 1151--1160.

\bibitem{Bhattad2020Unrestricted}
A.~Bhattad, M.~J. Chong, K.~Liang, B.~Li, and D.~A. Forsyth, ``Unrestricted
  adversarial examples via semantic manipulation,'' in \emph{ICLR}, 2020.

\bibitem{NEURIPS2018_8cea559c}
Y.~Song, R.~Shu, N.~Kushman, and S.~Ermon, ``Constructing unrestricted
  adversarial examples with generative models,'' in \emph{NeurIPS},
  vol.~31.\hskip 1em plus 0.5em minus 0.4em\relax Curran Associates, Inc.,
  2018.

\bibitem{gowal2020achieving}
S.~Gowal, C.~Qin, P.-S. Huang, T.~Cemgil, K.~Dvijotham, T.~Mann, and P.~Kohli,
  ``Achieving robustness in the wild via adversarial mixing with disentangled
  representations,'' in \emph{CVPR}, 2020, pp. 1211--1220.

\bibitem{qiu2020semanticadv}
H.~Qiu, C.~Xiao, L.~Yang, X.~Yan, H.~Lee, and B.~Li, ``Semanticadv: Generating
  adversarial examples via attribute-conditioned image editing,'' in
  \emph{ECCV}.\hskip 1em plus 0.5em minus 0.4em\relax Springer, 2020, pp.
  19--37.

\bibitem{wong2020learning}
E.~Wong and J.~Z. Kolter, ``Learning perturbation sets for robust machine
  learning,'' in \emph{ICLR}, 2021.

\bibitem{prabhu2018art}
V.~U. Prabhu, J.~Whaley, and S.~Francisco, ``Art-attack! on style transfers
  with textures, label categories and adversarial examples,'' 2018.

\bibitem{zhang2019theoretically}
H.~Zhang, Y.~Yu, J.~Jiao, E.~P. Xing, L.~E. Ghaoui, and M.~I. Jordan,
  ``Theoretically principled trade-off between robustness and accuracy,'' in
  \emph{ICML}, 2019.

\bibitem{zhang2021_GAIRAT}
J.~Zhang, J.~Zhu, G.~Niu, B.~Han, M.~Sugiyama, and M.~Kankanhalli,
  ``Geometry-aware instance-reweighted adversarial training,'' in \emph{ICLR},
  2021.

\bibitem{hitaj2021evaluating}
D.~Hitaj, G.~Pagnotta, I.~Masi, and L.~V. Mancini, ``Evaluating the robustness
  of geometry-aware instance-reweighted adversarial training,'' \emph{arXiv
  preprint arXiv:2103.01914}, 2021.

\bibitem{zhang2018unreasonable}
R.~Zhang, P.~Isola, A.~A. Efros, E.~Shechtman, and O.~Wang, ``The unreasonable
  effectiveness of deep features as a perceptual metric,'' in \emph{CVPR},
  2018, pp. 586--595.

\bibitem{cohen2019certified}
J.~Cohen, E.~Rosenfeld, and Z.~Kolter, ``Certified adversarial robustness via
  randomized smoothing,'' in \emph{ICML}.\hskip 1em plus 0.5em minus
  0.4em\relax PMLR, 2019, pp. 1310--1320.

\bibitem{10.5555/3454287.3455300}
H.~Salman, G.~Yang, J.~Li, P.~Zhang, H.~Zhang, I.~Razenshteyn, and S.~Bubeck,
  ``Provably robust deep learning via adversarially trained smoothed
  classifiers,'' in \emph{NeurIPS}.\hskip 1em plus 0.5em minus 0.4em\relax Red
  Hook, NY, USA: Curran Associates Inc., 2019.

\bibitem{zhang2020towards}
H.~Zhang, H.~Chen, C.~Xiao, S.~Gowal, R.~Stanforth, B.~Li, D.~Boning, and C.-J.
  Hsieh, ``Towards stable and efficient training of verifiably robust neural
  networks,'' in \emph{ICLR}, 2020.

\bibitem{leino21gloro}
K.~Leino, Z.~Wang, and M.~Fredrikson, ``Globally-robust neural networks,'' in
  \emph{ICML}, 2021.

\bibitem{huang2017arbitrary}
X.~Huang and S.~Belongie, ``Arbitrary style transfer in real-time with adaptive
  instance normalization,'' in \emph{ICCV}, 2017, pp. 1501--1510.

\bibitem{wang2004image}
Z.~Wang, A.~C. Bovik, H.~R. Sheikh, and E.~P. Simoncelli, ``Image quality
  assessment: from error visibility to structural similarity,'' \emph{IEEE
  transactions on image processing}, vol.~13, no.~4, pp. 600--612, 2004.

\bibitem{sharif2018suitability}
M.~Sharif, L.~Bauer, and M.~K. Reiter, ``On the suitability of lp-norms for
  creating and preventing adversarial examples,'' in \emph{Proceedings of the
  IEEE Conference on Computer Vision and Pattern Recognition Workshops}, 2018,
  pp. 1605--1613.

\bibitem{deng2009imagenet}
J.~Deng, W.~Dong, R.~Socher, L.-J. Li, K.~Li, and L.~Fei-Fei, ``{ImageNet}: A
  large-scale hierarchical image database,'' in \emph{CVPR}.\hskip 1em plus
  0.5em minus 0.4em\relax IEEE, 2009, pp. 248--255.

\bibitem{rw2019timm}
R.~Wightman, ``Pytorch image models,''
  \url{https://github.com/rwightman/pytorch-image-models}, 2019.

\bibitem{Wong2020Fast}
E.~Wong, L.~Rice, and J.~Z. Kolter, ``Fast is better than free: Revisiting
  adversarial training,'' in \emph{ICLR}, 2020.

\bibitem{2019arXiv190412843S}
A.~Shafahi, M.~Najibi, M.~A. Ghiasi, Z.~Xu, J.~Dickerson, C.~Studer, L.~S.
  Davis, G.~Taylor, and T.~Goldstein, ``Adversarial training for free!'' in
  \emph{NeurIPS}, vol.~32.\hskip 1em plus 0.5em minus 0.4em\relax Curran
  Associates, Inc., 2019.

\bibitem{xie2019feature}
C.~Xie, Y.~Wu, L.~v.~d. Maaten, A.~L. Yuille, and K.~He, ``Feature denoising
  for improving adversarial robustness,'' in \emph{CVPR}, 2019, pp. 501--509.

\bibitem{mao2021rethinking}
X.~Mao, G.~Qi, Y.~Chen, X.~Li, S.~Ye, Y.~He, and H.~Xue, ``Rethinking the
  design principles of robust vision transformer,'' \emph{arXiv preprint
  arXiv:2105.07926}, 2021.

\bibitem{hendrycks2021many}
D.~Hendrycks, S.~Basart, N.~Mu, S.~Kadavath, F.~Wang, E.~Dorundo, R.~Desai,
  T.~Zhu, S.~Parajuli, M.~Guo, D.~Song, J.~Steinhardt, and J.~Gilmer, ``The
  many faces of robustness: A critical analysis of out-of-distribution
  generalization,'' \emph{ICCV}, 2021.

\bibitem{xie2020self}
Q.~Xie, M.-T. Luong, E.~Hovy, and Q.~V. Le, ``Self-training with noisy student
  improves {ImageNet} classification,'' in \emph{CVPR}, 2020, pp.
  10\,687--10\,698.

\bibitem{liu2021Swin}
Z.~Liu, Y.~Lin, Y.~Cao, H.~Hu, Y.~Wei, Z.~Zhang, S.~Lin, and B.~Guo, ``Swin
  transformer: Hierarchical vision transformer using shifted windows,''
  \emph{arXiv preprint arXiv:2103.14030}, 2021.

\bibitem{NEURIPS2019_5d4ae76f}
F.~Tramer and D.~Boneh, ``Adversarial training and robustness for multiple
  perturbations,'' in \emph{NeurIPS}, vol.~32.\hskip 1em plus 0.5em minus
  0.4em\relax Curran Associates, Inc., 2019.

\bibitem{recht2019imagenet}
B.~Recht, R.~Roelofs, L.~Schmidt, and V.~Shankar, ``Do {ImageNet} classifiers
  generalize to {ImageNet}?'' in \emph{ICML}.\hskip 1em plus 0.5em minus
  0.4em\relax PMLR, 2019, pp. 5389--5400.

\bibitem{taori2020measuring}
R.~Taori, A.~Dave, V.~Shankar, N.~Carlini, B.~Recht, and L.~Schmidt,
  ``Measuring robustness to natural distribution shifts in image
  classification,'' in \emph{NeurIPS}, 2020.

\bibitem{geirhos2018imagenet}
R.~Geirhos, P.~Rubisch, C.~Michaelis, M.~Bethge, F.~A. Wichmann, and
  W.~Brendel, ``{ImageNet}-trained cnns are biased towards texture; increasing
  shape bias improves accuracy and robustness,'' \emph{arXiv preprint
  arXiv:1811.12231}, 2018.

\end{thebibliography}


\newpage
\onecolumn
\section{Appendix}
\subsection{More Visualization Results}
\label{Visualization}

\end{document}